\documentclass[sigconf]{acmart}

\usepackage[para]{footmisc}

\usepackage{amsmath}
\usepackage{natbib}
\usepackage{amsfonts}
\usepackage{graphicx}
\usepackage{hhline}
\usepackage{hyperref}
\usepackage{enumerate}
\usepackage{multirow}
\usepackage{array}
\usepackage[para]{footmisc}
\usepackage{adjustbox}
\usepackage{balance}
\usepackage{subcaption}
\usepackage[normalem]{ulem}
\usepackage[show]{chato-notes}
\usepackage{marginnote}

\usepackage{subcaption}

\usepackage{algorithm}
\usepackage{algorithmicx}
\usepackage{algpseudocode}

\usepackage{enumitem}

\newcommand{\iadh}[1]{\textcolor{black}{#1}}

\newcommand{\zixuan}[1]{\textcolor{black}{#1}}
\newcommand{\yi}[1]{\textcolor{black}{#1}}
\newcommand{\io}[1]{\textcolor{black}{#1}} 
\newcommand{\cm}[1]{\textcolor{black}{#1}}
\newcommand{\zx}[1]{\textcolor{black}{#1}}
\newcommand{\ya}[1]{\textcolor{black}{#1}}
\newcommand{\yiz}[1]{\textcolor{black}{#1}}
\newcommand{\yay}[1]{\textcolor{black}{#1}}
\newcommand{\yy}[1]{\textcolor{black}{#1}}

\begin{document}

\title{A Unified Graph Transformer for Overcoming Isolations in Multi-modal Recommendation }
% \title{UGT: A Unified Multi-modal Graph Transformer for Recommendation}

% \copyrightyear{2024} 
% \acmYear{2024} 
% \setcopyright{acmlicensed}\acmConference[SIGIR '24]{The Web Conference}{May 13--17, 2024}{Singapore, Singapore}
% \acmBooktitle{The Web Conference), May 13--27, 2024, Singapore, Singapore}

\copyrightyear{2024}
\acmYear{2024}
\setcopyright{acmcopyright}
% \acmConference[RecSys '24]{Proceedings of the 47th International ACM RecSys Conference}{July 14--18, 2024}{Bari, Italy)}
% \acmBooktitle{Proceedings of the 47th International ACM SIGIR Conference on Research and Development in Information Retrieval (SIGIR '24), July 14--18, 2024, Washington D.C., USA}

\author{Zixuan Yi}
\email{z.yi.1@research.gla.ac.uk}
\affiliation{%
 \institution{University of Glasgow}
 \city{Glasgow}
 \state{Scotland}
 \country{United Kingdom}
}

\author{Iadh Ounis}
\email{iadh.ounis@glasgow.ac.uk}
\affiliation{%
 \institution{University of Glasgow}
 \city{Glasgow}
 \state{Scotland}
 \country{United Kingdom}
}

% \author{Craig Macdonald}
% \email{craig.macdonald@glasgow.ac.uk}
% \affiliation{%
%  \institution{University of Glasgow}
%  \city{Glasgow}
%  \state{Scotland}
%  \country{United Kingdom}
% }

% \author{Anonymous name}
% \affiliation{
% \institution{Example Organisation \\ email@example.com}
% \country{Anonymous country}
% }

\fancyhead{}
\begin{abstract}
\looseness -1 \zixuan{With the rapid development of online multimedia services, especially in e-commerce platforms, there \iadh{is} a pressing need for personalised \yy{recommender} systems that can effectively encode the diverse \iadh{multi-modal} content associated with each item.}
% Due to the success of the Collaborative Filtering (CF) method, multi-modal recommender systems exploit multi-modal information to obtain more accurate users’ preferences on top of the CF paradigm.
However, \zixuan{we argue that} existing multi-modal recommender systems \iadh{typically use isolated \ya{processes for both}} feature extraction and 
\zx{modality \yy{encoding}. \ya{Such isolated processes can harm the recommendation performance.}}
% multiple independent processing for each modality. 
Firstly, \iadh{an} isolated extraction \iadh{process} underestimates the importance of effective feature extraction in multi-modal recommendations, potentially incorporating \iadh{non-relevant} information, \iadh{which is} harmful to item representations. 
\ya{Second, an} \zx{isolated modality encoding process} produces \yay{disjoint} embeddings for item modalities \zx{due to the individual processing of each modality}, which leads \zixuan{to} a suboptimal fusion of user/item representations for \yay{an} \iadh{effective} user preferences prediction. 
\zx{We hypothesise that the use of a unified model \ya{for addressing both aforementioned isolated processes} will enable the consistent extraction and cohesive fusion of joint multi-modal features, thereby enhancing the effectiveness of multi-modal recommender systems.}
% We hypothesise that by employing a unified model for multi-modal representation learning, we can uniformly extract and fuse \iadh{the joint} multi-modal features \iadh{for effective} multi-modal recommendation. 
\iadh{In this paper, we} propose a novel model, \zixuan{called} Unified \yy{m}ulti-modal Graph Transformer (UGT), \iadh{which} \yi{firstly} leverages a multi-way transformer to extract aligned multi-modal features from raw data \yi{for top-$k$ recommendation}. 
% \inote{thereby?} modelling \iadh{the users' preferences} with 
Subsequently, 
\zx{we {build}}
% our UGT \io{model} employs
a unified graph neural network \zx{in our UGT \io{model}}
to jointly \zx{fuse} \zixuan{the} \yiz{multi-modal user/item representations derived from the \yay{output of the} multi-way transformer.} 
% user/item representations \zixuan{with their corresponding multi-modal features}.
\zx{Using the graph transformer architecture of our UGT model,}
% Based on \zixuan{\zx{our devised} graph transformer architecture},
% \inote{confusing, which one, you devise two things, UGT and a unified grah, whiuch you mean in this sentence?}, 
we show that \zx{the UGT model} achieve\yy{s} significant effectiveness gains, especially when jointly optimised with the \yay{commonly used} 
% multi-modal 
recommendation \yi{losses}.
Our extensive experiments conducted on three benchmark datasets demonstrate the superiority of our proposed UGT model over \yiz{nine} existing state-of-the-art recommendation approaches. \looseness -1
% \pageenlarge{4}
\end{abstract}


% \ccsdesc[500]{Computer systems organization~Embedded systems}
% \ccsdesc[300]{Computer systems organization~Redundancy}
% \ccsdesc{Computer systems organization~Robotics}
% \ccsdesc[100]{Networks~Network reliability}

% \keywords{Multi-modal; Self-supervised Learning; Graph Neural Network}

\maketitle

\section{Introduction}\label{sec:intro}
% \pageenlarge{1}

% \begin{figure}[tb]
%     \centering
%     \resizebox{0.48\textwidth}{!}{
%         \includegraphics[trim={1.5cm 11cm 1.5cm 11cm},clip]{fig_problem.pdf}
%     }
%     \caption{Fig.1: Recommendation working flow in exiting multi-modal recommendation models.}
%     \label{fig:prob}
%     % \vspace{-4mm}
% \end{figure}

In specific domains such as fashion~\cite{wu2022multimodal}, music~\cite{park2022exploiting}, and micro-video recommendation~\cite{yi2022multi}, \iadh{the effectiveness of} recommender systems \iadh{can be improved} \zixuan{by using}
% be effectively supported in their decision-making process by 
% all types of 
multi-modal data sources \iadh{that} the users usually interact with (e.g., product images and descriptions, users’ reviews, audio tracks).
\iadh{Indeed, several previous} studies~\cite{wang2023large,liu2023multimodal,zhou2023comprehensive} \zixuan{have proposed} multi-modal recommender systems that leverage multi-modal (i.e., audio, visual, textual) content data to augment the representation of items, \iadh{so as to} tackle \yay{the problem of data sparsity} in the user-item matrix.
% and the inexplicable nature of users’ actions (e.g., clicks, views) on online platforms which may not always be easy to profile for the recommendation algorithms.

\begin{figure}[tb]
        \includegraphics[trim={1.5cm 11cm 1.5cm 11cm},clip,width=1\linewidth]{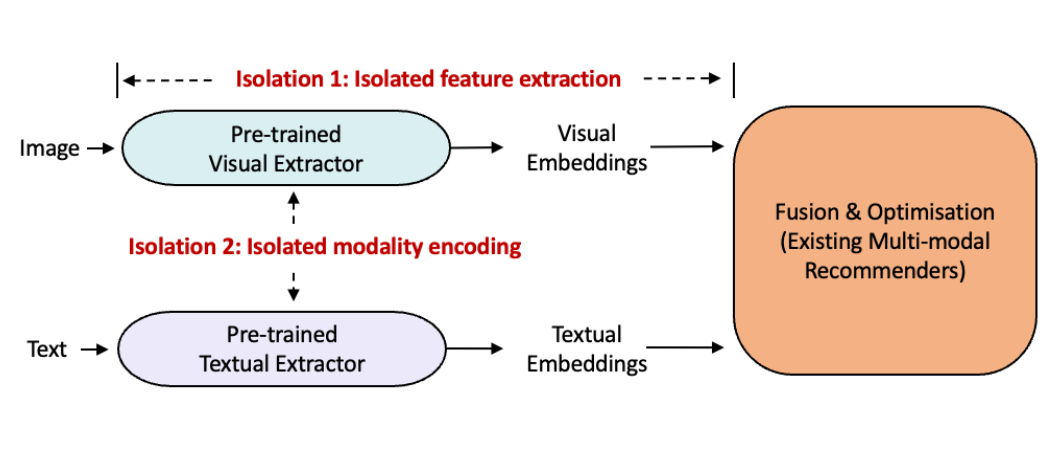}\vspace{-2em}
        \captionsetup{labelformat=empty}
        \caption{Fig.1: Recommendation working flow in exiting multi-modal recommendation models.}\label{fig:prob}
 \vspace{-5mm}
\end{figure}

% The --- of multi-modal data not only contributes to mitigating these issues but also unlocks the potential to understand and represent items comprehensively,
%\zixuan{The \iadh{coherent} use} of multi-modal data has garnered attention and emphasis in recent research~\cite{liu2023multimodal,deldjoo2020recommender}, highlighting its importance in enhancing multi-modal recommendation systems.
% \pageenlarge{4}
\iadh{Recent research~\cite{liu2023multimodal,deldjoo2020recommender}  has shown the importance and benefits of coherently using multi-modal data in enhancing multi-modal \yy{recommender} systems.} 
\zixuan{\iadh{Multi-modal} feature extraction, which aims to derive meaningful patterns and information from multiple types of data,} is fundamental to the initial stage of any multi-modal recommendation pipeline. \iadh{Indeed, effectively} performing \iadh{such feature} extraction is critical \yay{for ensuring} high-quality \yay{recommendations}~\cite{liu2023multimodal,zhou2023comprehensive}.
% \zixuan{Although} the multi-modal feature extraction is fundamental to the initial stage of any multi-modal recommendation pipeline,  extracting effective multi-modal features is crucial in providing high-quality recommendations~\cite{liu2023multimodal,zhou2023comprehensive}. 
\zx{On the other hand, multi-modal fusion \yay{is also crucial} \yay{for creating} user/item representations from \yay{various} \ya{contextual} data, \yay{including} \ya{the items'} images and descriptions. \ya{Such a} fusion typically \ya{occurs} after feature extraction. 
%involves 
% an integrated 
%a fusion component.
\yiz{\yay{In fact}, most existing multi-modal recommendation models primarily function as a fusion component,}
designed to merge the extracted features into a joint representation \zx{of the items}~\cite{zhou2023comprehensive}.}
\zx{\yi{Figure~\ref{fig:prob} \ya{illustrates} \io{\ya{how} multi-modal recommendation} typically \ya{consists of a} \io{pipeline of workflows ranging} from data ingestion \io{and} feature extraction, \io{to} multi-modal fusion and optimisation}.}
% for user/item modelling
\zx{Within this pipeline, two important problems emerge, \iadh{namely, the use of (i) \ya{an} isolated feature extraction \ya{process} (c.f. Isolation 1 \yay{in the figure}) and  
(ii) \zx{\ya{an} isolated modality encoding process (c.f. Isolation 2)}.}}
% \zx{These \iadh{two problems are} visualised \iadh{and denoted as} `Isolation 1' and `Isolation 2' in Figure~\ref{fig:prob}.}
% \zixuan{In multi-modal recommendation, particularly in the context of using multi-modal data, two important problems emerge, \iadh{namely, the use of (i) \ya{an} isolated feature extraction \ya{process} and  
% (ii) \zx{\ya{an} isolated modality modelling process}. }
% \yi{multiple} independent processing for each modality. 
% \iadh{Such problems pose challenges to the multi-modal recommendation system, and limit its} optimal functionality.
% The \iadh{two problems are} visualised \iadh{and denoted as} `Isolation 1' and `Isolation 2' in Figure~\ref{fig:prob}. 
\iadh{\yay{These two problems}} impede the system from \yay{exploring} the full potential of \io{the} multi-modal data by creating isolated processes,
\yi{as they ignore the collaborative power of unifying various modalities and leveraging an integrated architecture that effectively combines} 
\io{the} extraction process with \yay{the} fusion process.
% \io{the} extraction component with a fusion component.
% \inote{unclear what this fusion component is and what it is for, and how it relates to one of the isolation problems; Fig 1 does not say what is being fusioned and why}.
% integrated 
% unifying various modalities and leveraging an integrated architecture that effectively combines an extraction module with a fusion module.
% unified modalities and architecture \inote{unclear, what architecture?}.
% \yi{As illustrated in Figure~\ref{fig:prob}, \io{multi-modal recommendation} \io{consists of a pipeline of workflows ranging} from data ingestion \io{and} feature extraction, \io{to} multi-modal fusion and optimisation.}
\zixuan{\iadh{The} `Isolation~1' \iadh{problem} \io{refers to} the challenges arising from 
employing \textit{an isolated \zx{feature} extraction process} \yi{within each workflow of the multi-modal \io{recommendation} system.}}
% in each recommendation framework 
% \inote{unclear, since a recsys has a single recommendation process, unless you want to say there is a recsys framework for each modality that need later to be fusioned.}
% poses challenges. 
% Firstly, 
\zixuan{Specifically,} the pre-trained \yiz{feature} extractors 
% for each modality 
function independently 
\yiz{alongside the subsequent fusion component.}
% within \yi{the multi-modal recommendation \yi{pipeline}.}
% \inote{there is only ONE recommendation framework in a given system}.
% the individual pre-trained extractors, used for each modality in each recommendation framework, operate independently from the recommendation model. 
% This individual processing of each modality 
This isolated extraction process
potentially \io{leads to the incorporation of} non-relevant information to the subsequent fusion \zx{component},
% \inote{is that the same as what you called fusion component (singular) above? why it is now plural, in which sense it is a model, unclear what it/they fusion(s)},}
% This setup ignores the opportunity to jointly optimise them alongside subsequent fusion models \inote{I cannot parse this sentence; simplify and keep things simple},  
\ya{thereby} \yi{impeding the extraction of} more effective multi-modal features.
% \inote{the last bit 'bypassing ...' is not clear; is it necessary?}.
% Secondly, 
% Moreover, \zixuan{the} diverse implementations \inote{what does it mean? why each extractor has a different implementation?} of extractors \zixuan{also} complicate \iadh{a} fair comparison \inote{between what and what? what has this anything to do with reproducibility?} and reproducibility \zixuan{among various multi-modal recommendation frameworks \inote{I do not get why there are many frameworks; can you say what each of these framework do?}.}
% impede the interdependence across various multi-modal recommendation frameworks, 
% Secondly, despite the availability of numerous pre-trained deep learning models in popular open-source libraries, the lack of shared interfaces for feature extraction across them represents a challenge for reproducibility~\cite{wei2019mmgcn}.
On the other hand, as indicated \iadh{by the} `Isolation 2' \iadh{problem} in Figure~\ref{fig:prob}, the existing 
% multi-modal 
recommender systems
% apply multiple-stream methods to 
process each modality separately \iadh{before} finally \iadh{fusing} them with simple concatenations~\cite{yi2022multi,wei2019mmgcn}. \iadh{Such \textit{\zx{an isolated modality encoding process}} misses} opportunities to jointly optimise the user/item embeddings \iadh{resulting} from \zixuan{the different} modalities.
\zx{To address \yay{these} isolated processes for both feature extraction \yay{(Isolation 1)} and modality modelling \yay{(Isolation 2)},}
% \yi{To address the aforementioned problems \inote{again which ones?},}
we introduce a Unified Graph Transformer (UGT) \yi{model for top-$k$ multi-modal recommendation.} \yay{UGT combines a multi-way transformer with a newly designed unified Graph Neural Network (GNN) in a novel cascading graph transformer architecture. We use the multi-way transformer as the extraction component and integrate
its output in the \yiz{unified} GNN that serves as the fusion component, so as to address the ‘Isolation 1’ problem}.
%\zx{Specifically, we combine a multi-way transformer and \ya{a} newly designed unified Graph Neural Network (GNN) in a cascading graph transformer architecture to solve \ya{both} the isolated feature extraction
% (refer to Isolation 1) 
%and modality modelling 
% (refer to Isolation 2) 
%problems in multi-modal recommendation.}
%To address the `Isolation 1' problem, we leverage \ya{a} multi-way transformer as the extraction component 
%\yiz{and integrate its output in a newly devised GNN that serves as the fusion component.}
% and closely pair it with a GNN as the fusion component.
\yay{This new unified GNN within our proposed UGT architecture allows to}
\zixuan{address \io{the} `Isolation 2' \io{problem}} \yay{by}
%we \zx{devise} a \ya{new} unified GNN \zx{within our UGT architecture} that 
uniformly \yay{aggregating} the users/items' neighbour information and their corresponding multi-modal features, thereby \yiz{enriching} \yi{the} multi-modal fusion for \yiz{\yay{better} user/item representations}.
% \yi{the} user/item modelling.
\yiz{The integration of the multi-way transformer and the unified GNN enables a simultaneous optimisation \ya{of} the extraction of effective multi-modal features and their subsequent fusion, thereby solving the isolated feature extraction process.}
% \inote{This text breaks the flow, we need isolation 2 first} \yi{\ya{Such an} integration enables \io{a} simultaneous optimisation \ya{of} the extraction of effective multi-modal features and their subsequent fusion, thereby enhancing \iadh{the} final user/item embeddings.}
\zx{\ya{\yay{Our} work is} the first to use \ya{a} multi-way transformer as an extraction component in multi-modal recommendation, \ya{thereby going beyond its current use \yiz{in the literature} \yay{for encoding} various modalities (e.g., images, text) within a shared transformer block~\cite{bao2022vlmo}} in Image-Text retrieval \yiz{or} Visual Question Answering (VQA)~\cite{wang2023image,ge2021structured,ge20243shnet}.}
% \zx{To the best of our knowledge, we are the first to use the multi-way transformer as an extraction component in multi-modal recommendation, which is typically used in Image-Text retrieval and Visual Question Answering (VQA)~\cite{bao2022vlmo} to encode various modalities (e.g., images, text) within a shared transformer block.}
\yiz{\yiz{Indeed, we} \yiz{instead} apply \yiz{the multi-way transformer} in a recommendation setting} to transform heterogeneous data into a unified latent space, thereby producing aligned embeddings from each modality.
% \inote{so you cite 29 but if 29 has done in recommendation, you are saying it is not novel; if 29 is not a recommendation paper, what you are citing 29 for?}.
%\yiz{We then use our unified GNN to uniformly fuse these aligned multi-modal embeddings, rather than processing each modality individually.}
\yay{We then use our proposed unified GNN \yiz{component} to seamlessly \yiz{fuse} the resulting aligned multi-modal embeddings, a significant departure from the existing models that process modalities separately.}

\looseness -1
In summary, our contributions \io{in this paper} are three-folds: 
$(1)$~\zx{We \zx{propose} a novel \zx{Unified Graph Transformer (UGT)} model to resolve}
% We \yi{integrate} a multi-way transformer network \yi{as the extraction component and pair it with a unified GNN as the fusion component,} 
% % seamlessly extract \inote{what does it mean? how you measure semlessness?} multi-modal features from raw data, 
% thereby \zixuan{resolving} 
the isolated feature extraction problem in multi-modal recommendation. 
To \zixuan{the} best of our knowledge, \yiz{UGT is} the first \yi{end-to-end} multi-modal recommendation model \yi{that directly facilitates multi-modal recommendation from raw data.} 
% \inote{INCOMPLETE SENTENCE}.
% \iadh{which} provides an end-to-end solution \inote{to what? why all these unecessary words? why not say you propose and end-to-end model to do X}. 
Our \yiz{UGT model} not only extracts  \yi{the} \zixuan{items' features} and integrates them with \yi{the} user-item interactions but also simultaneously optimises extraction and fusion under a unified \yi{architecture;} 
% \inote{STOP USING UNECESSARY NEW CONCEPTS - WHAT IS AN OBJECTIVE FRAMEWORK? IS THAT THE SAME AS THE UNIFIED ONE YOU HAVE BEEN TALKING ABOUT FOR @ COLUMNS?}; 
$(2)$ We introduce a \zx{new} unified GNN \zx{as the fusion component} to uniformly fuse the user-item interactions and their multi-modal features in a unified manner. 
This \yi{unified} approach \io{addresses} the \ya{problem of} 
\zx{isolated modality encoding}
% \yi{multiple} independent processing
% tradition  \inote{what tradition, people have traditions, not systems}  of individual modalities 
in the existing \io{multi-modal} recommender systems, \yi{\yiz{so as} to enhance} the recommendation performance by ensuring \io{a} more effective modelling of the target users \zx{from \ya{the} items' images and descriptions}; 
% integrated feature fusion and interaction modelling.
% disrupts the conventional multi-stream processing tradition of individual modalities in the existing recommender systems and improving recommendation performance.
% uniformly fusing image and text tokens in a joint manner;
% a cascading graph transformer architecture, initially employing a multi-way transformer network followed by a unified graph neural network.
% This approach disrupts the conventional multi-stream processing of individual modalities, uniformly fusing image and text tokens in a joint manner;
% We propose a first multi-way transformer network then unified graph neural network cascading graph transformer architecture, which breaks the tradition of multi-stream processing of individual modalities and uniformly fuses the image/text tokens in a shared/joint manner;
$(3)$ We conduct extensive experiments on three public datasets to demonstrate \iadh{the} effectiveness of our UGT model in comparison to \yiz{nine} \yay{existing} state-of-the-art multi-modal recommender systems. Moreover, we provide visualisations \yiz{and measure the average Mean Square Error between the visual and textual embeddings to show} that \yiz{UGT successfully aligns these embeddings to a unified and \yay{tighter} semantic space.} \looseness -1

\section{Related Work}\label{sec:rwork}
In this section, \iadh{we position our work in relation to} related methods and techniques \iadh{in the literature}, namely multi-modal recommendation \yay{(Section~\ref{sec:mmrec})} and graph transformer\yi{s} \yay{(Section~\ref{sec:gt})}.

% \vspace{-2mm}
% \pageenlarge{1}
\subsection{Multi-modal Recommendation}\label{sec:mmrec}
% modal and extraction
% cf content-based
% separated extraction and fusion

% In terms of multi-modal recommender systems, existing models either incorporate multi-modal item data into the traditional Collaborative Filtering (CF) paradigm using deep or graph learning techniques combined with feature extraction methods, or they leverage pre-extracted features~\cite{liu2023multimodal,zhou2023comprehensive}. These models exploit item images and descriptions to enrich the recommendation systems, 
% % implementing a variety of techniques to effectively enhance recommendation performance.
% applying diverse techniques to effectively enhance recommendation performance.

\looseness -1 \zixuan{Existing} multi-modal \iadh{recommendation models} incorporate multi-modal \yay{data sources} \yay{for items} into the traditional Collaborative Filtering (CF) paradigm through the application of deep or graph learning techniques alongside different feature extraction methods~\cite{liu2023multimodal,zhou2023comprehensive}. 
\zixuan{All such} models capitalise on \iadh{the items'} multi-modalities to enhance the recommendation \iadh{performance}.
% In terms of multi-modal recommender systems, existing models utilise deep or graph learning techniques, combined with feature extraction methods, to incorporate multi-modal item data into the traditional Collaborative Filtering (CF) paradigm~\cite{liu2023multimodal,zhou2023comprehensive}. 
% These models exploit multi-modalities of items to enrich the recommendation systems, 
% % implementing a variety of techniques to effectively enhance recommendation performance.
% applying diverse techniques to effectively enhance recommendation performance.
For instance, VBPR~\cite{he2016vbpr} enhanced the BPR method~\yiz{\cite{rendle2012bpr}} by modelling \iadh{the} users' preferences \iadh{in relation to} visual information and extract\zx{ed} visual features from \iadh{the items'} images through a pre-trained Deep Convolutional Neural Network (CNN)~\cite{donahue2014decaf}. 
Moreover, MMGCN~\cite{wei2019mmgcn} used Graph Convolutional Networks (GCN) to propagate modality-specific embeddings and \iadh{to} capture \iadh{the} users' preferences related to these modalities. 
For feature extraction, MMGCN employed \zixuan{a variant of \yay{the} Residual Network model} (ResNet50~\cite{he2016deep}) for visual features, Sentence2Vector~\cite{le2014distributed} for textual features, and VGGish~\cite{gemmeke2017audio} for acoustic features. 
MMGCL~\cite{yi2022multi}, SLMRec~\cite{tao2022self} \yiz{and BM3~\cite{zhou2023bootstrap}} followed the same feature extraction methods \zixuan{as MMGCN.}
\zixuan{However,} MMGCL \yay{proposed} a self-supervised graph learning model that leverages \yi{modality-based} edge dropout and modality masking to \yi{capture the} complex user preferences. 
\iadh{On the other hand}, SLMRec constructed fine and coarse spaces to align features across modalities,
% merely applying 
\yiz{then merely applied} a contrastive learning loss \yiz{for} multi-modal recommendation.
\yiz{BM3 bootstrapped the user/item node embeddings \yiz{through a} dropout augmentation and reconstructed the user-item graph for \yiz{multi-modal recommendation}.} \yay{In a different line of work,}
LATTICE~\cite{LATTICE21} \yiz{built} item-item graphs from multi-modal features and \yay{uncovered} latent semantic structures between items. \yay{Later,} \yiz{ FREEDOM~\yiz{\cite{zhou2023tale}} \yay{further} improved LATTICE by pruning popular edges on the item-item graphs.}
\yiz{Both LATTICE and FREEDOM} used Deep CNN~\cite{donahue2014decaf} and Sentence-Transformer~\cite{reimers2019sentence} for visual and textual feature extraction, respectively.
\ya{While} each multi-modal \iadh{recommendation model} leverages features extracted from pre-trained \yy{extraction} models and employs specific strategies like Graph Neural Network (GNN) for multi-modal fusion, these \zx{recommendation models}
primarily rely on pre-trained extraction models for \yiz{feature extraction.}
% each modality. 
As a result, the use of pre-trained models for extraction isolate\yy{s} the extraction step from the \yi{entire}
process of multi-modal recommendation.
\yi{This isolation occurs because the extraction models are not tailored or fine-tuned for the multi-modal recommendation task,}
thus 
\zx{potentially incorporating \iadh{non-relevant} information that could harm the \yi{item representations} 
in a multi-modal recommendation context.}
In contrast to previous \yi{multi-modal recommendation models} that \zx{use separate extraction models}, \yay{in this paper,} we propose \yay{instead} a \yay{novel} unified graph transformer model that 
seamlessly integrates the extraction and fusion \yay{processes} 
\yiz{by combining a multi-way transformer with a unified GNN in a cascading architecture},
% \inote{by doing what new things exactly? by using what new things (components, layers, etc)? viz models you describe above}, 
thereby overcoming the isolated extraction process \yy{(Isolation 1)} in multi-modal recommendation. \looseness -1

\vspace{-2mm}
% \pageenlarge{4}
\subsection{Graph Transformers}\label{sec:gt}

\looseness -1 The integration of transformer architectures into graph neural networks has been proposed as a promising solution to enhance existing Graph Neural Networks (GNNs)~\cite{ying2021transformers,yi2023graph}.
In \yay{the} recommendation \yiz{domain}, taking advantage of \yay{because of its} contextualised information modelling, the transformer network recently \yi{drew} much research attention from the \yay{recommendation research community} to explore the \yi{relationships} among items.
For instance, prior \yi{works}~\cite{li2023edge,cao2022contrastive} \yi{leveraged GNNs to model the item \yay{embeddings} within a user's sequence of items,}
% \iadh{constructed} \yi{an inter-sequence graph } the user-item interactions as the \yi{a} sequential structure 
\yi{subsequently integrating them} into transformer-based recommendation \yay{models} \yi{(e.g. SASRec~\cite{kang2018self}, Bert4Rec~\cite{sun2019bert4rec})} \iadh{in order} to \zixuan{better} model user behaviour sequences. 
% \inote{this is very unclear, there is no related work description whatsoever, it is not even english}
% first utilize a graph neural network
% to mine inter-sequence item collaborative relationship, and then
% exploit sequential attentive encoder
% Different from these sequential recommendation models. 
Moreover, existing approaches~\yiz{\cite{li2023graph, zhu2022graph1, chen2024sigformer,yi2023large,yi2023contrastive,yi2024directional}}
% \inote{only two? say e.g. then and ensure you cite the very last recsys papers on the topic} 
also \yi{integrated} \yi{the vanilla} transformer \yi{with} a GNN to exploit the global user-item relationship\yiz{s}. \yay{Such approaches enhance}
\yiz{the recommendation performance by \yay{going} beyond \yay{merely aggregating} local neighbours to \yay{encompassing} a global context.}
% instead of locally \yi{aggregating} neighbours information~\cite{li2023graph,xia2022self} \inote{say how this helps recommendation, otherwise it is a very shallow description}.
Different from these sequential and general recommendation models, the recent graph transformer\yiz{s} in multi-modal recommendation~\cite{liu2021pre1,wei2023lightgt} explicitly model\zx{led} the correlation between the item representations by jointly extracting \iadh{the} items' images and descriptions as node features.
Specifically, PMGT~\cite{liu2021pre1} 
constructed an item-item graph from pre-extracted multi-modal features and \yiz{used} a transformer network to model the item embeddings with their corresponding neighbours.
% pre-trained multi-modal graph transformer model to learn item
% representations individually by considering both \yi{the} item side information and their relationships.
LightGT~\cite{wei2023lightgt} \ya{proposed} a \zx{simplified graph} transformer architecture, \ya{featuring a} single-head attention to distil each type of pre-extracted multi-modal \ya{features}. 
% \inote{only 2 relevant moidels, no more recent ones?}
% and cascade \inote{unclear, split sentence} the transformer with a LightGCN~\cite{he2020lightgcn} model.
% to enhance the multi-modal recommendation performance.
\zx{Different from the aforementioned multi-modal graph transformers that process each modality individually, we introduce a new unified GNN as \yay{a} fusion component. This \yiz{fusion} component is closely paired with a multi-way transformer in our UGT architecture \yiz{so as to effectively fuse the extracted multi-modal features from the multi-way transformer}.}
% Different from the aforementioned multi-modal graph transformers, which process each modality individually, 
% In our proposed model, we leverage a multi-way transformer to explicitly extract aligned multi-modal features from raw data and devise a graph transformer architecture for multi-modal recommendation. 
% \yi{we devise a \ya{new} unified GNN that is paired with a multi-way transformer.}
% This combination \yiz{of the multi-way transformer and the unified GNN} uniformly aggregates the \yay{user-item} interactions and their joint multi-modal features\footnote{Although our UGT model primarily focuses on visual and textual modalities, we \iadh{would like to note that} it can generalise to other modalities \io{when they are present in the datasets}.}, \yi{thereby
% addressing the  \ya{problem of} \zx{isolated modality encoding} in multi-modal recommendation.}
In particular, we incorporate \yiz{a novel} attentive-fusion component 
% \inote{is that new or novel?} 
within the unified GNN to uniformly fuse the \yy{resulting multi-modal features from the multi-way transformer},
% into the aforementioned joint multi-modal features,  
% further enhancing the user/item representation learning 
\yy{thereby addressing the problem of isolated modality encoding \yy{(Isolation 2)}}
\ya{in} multi-modal recommendation.

\section{Model Architecture}\label{sec:meth}

\begin{figure}[tb]
    % \begin{subfigure}[t]{0.98\linewidth}
        \includegraphics[trim={1.5cm 7cm 1.5cm 6.5cm},clip,width=1\linewidth]{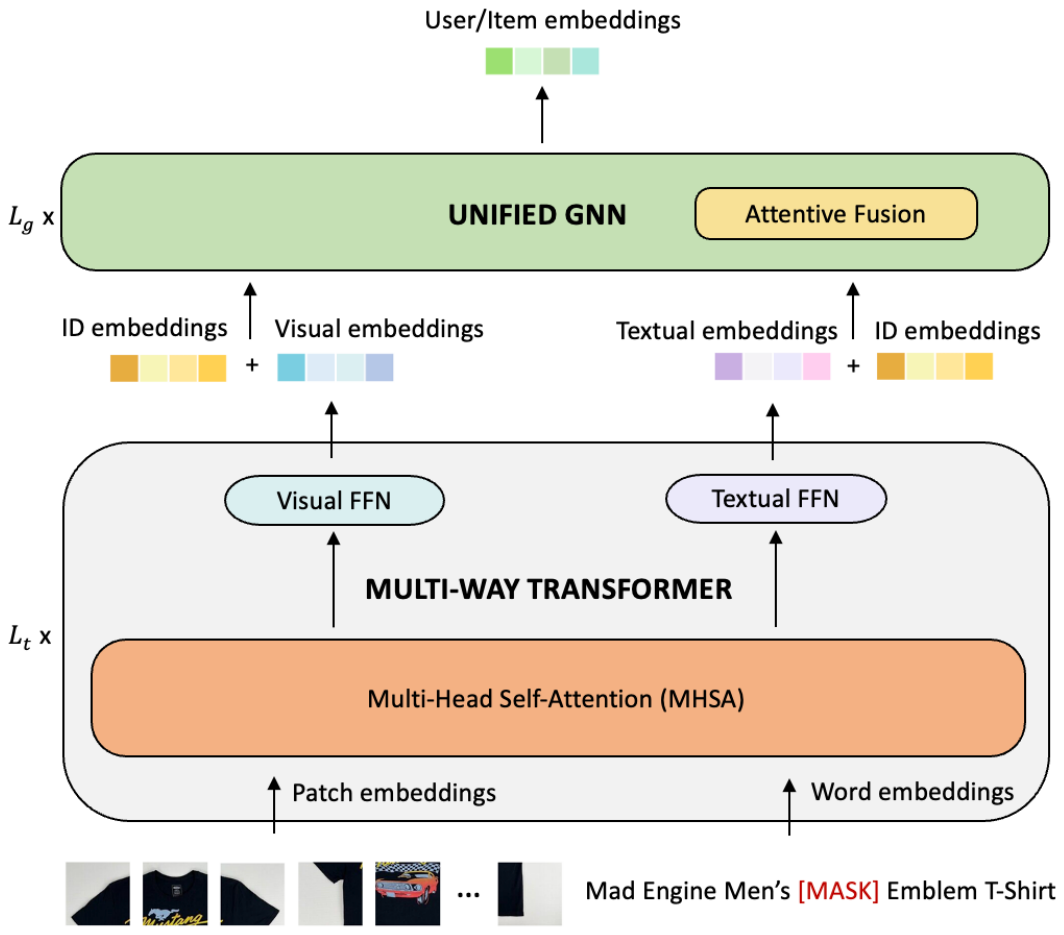}
        \captionsetup{labelformat=empty}
        \caption{Fig.2: \cm{Our} Unified multi-modal Graph Transformer (UGT).}
        \label{fig:arc}
    % \end{subfigure}
    \captionsetup{labelformat=simple}
\vspace{-4mm}
\end{figure}
In this section, we \yay{describe} our UGT model for top-$k$ multi-modal recommendation.
% \inote{1st time you mention top-k recommendation}. 
We first introduce the architecture of our UGT model in Section~\ref{sec:arc}, \iadh{which is illustrated} in Figure~\ref{fig:arc}.
\yay{Section~\ref{sec:mmencoding} presents} the unified multi-modal encoding process \yiz{of} our \yi{proposed} multi-way transformer.
Then, in Section~\ref{sec:ugnn}, we introduce the unified GNN, which explicitly fuses multi-modal features and user-item interactions into final user/item representations.
In Section~\ref{sec:obj}, we describe the optimisation of UGT with its corresponding losses. \looseness -1
% \inote{plural? it wasp lural then became singular then now plural again}.

\subsection{Preliminaries}\label{sec:prob}
Let $\mathcal{U}=\{u\}$ \yay{denote} the user set and $I=\{i\}$ \yay{denote} the item set. The input ID embeddings of user $u$ and item $i$ are $\mathrm{E}_{i d} \in \mathbb{R}^{d \times(|U|+|I|)}$. $d$ is the  dimension \yiz{of the user/item embedding}. 
Then, we denote each item modality \iadh{feature} as $\mathbf{E}_{i, m} \in \mathbb{R}^{d_m \times|I|}$, where $d_m$ is the dimension of \cm{that modality's \yay{features}}, $m \in \mathcal{M}$ is a modality, and $\mathcal{M}$ is the set of modalities. 
In this paper, we mainly consider visual and textual modalities \yiz{because the datasets used only contain these two types of raw data},
% \inote{WHY?}, 
\yay{hence} $\mathcal{M}=\{v, t\}$. 
\iadh{The} \yay{user's} historical behaviour data is denoted \yay{by} $\mathcal{R} \in\{0,1\}^{|U| \times|I|}$, where each entry $\mathcal{R}_{u, i}=1$ if user $u$ clicked item $i$, otherwise $R_{u, i}=0$. 
The historical interaction data $\mathcal{R}$ can be \iadh{seen} as a sparse behaviour graph $\mathcal{G}=\{\mathcal{V}, \mathcal{E}\}$, where $\mathcal{V}=\{\mathcal{U} \cup I\}$ denotes the set of nodes and $\mathcal{E}=\left\{(u, i) \mid u \in \mathcal{U}, i \in I, R_{u i}=1\right\}$ denotes the set of edges. The purpose of the \yi{multi-modal} recommendation \yiz{model} is to accurately predict \iadh{the} users' preferences by ranking items for each user according to \iadh{the} predicted preference scores $\hat{y}_{u i}$.

% \pageenlarge{4}
% \vspace{-2mm}
\subsection{Model Overview}\label{sec:arc}
Figure~\ref{fig:arc} \yay{shows} the model architecture of UGT, our \yay{proposed} graph transformer model,
\yi{which is composed of two \io{main} components: an extraction component (i.e., \cm{a} multi-way transformer) and a fusion component (i.e., \cm{a} unified GNN)}.
% The model architecture of UGT is illustrated in Figure~\ref{fig:arc}. 
Given \iadh{an} item image, \iadh{an} item text and \iadh{an} item ID, 
\yi{we \yiz{follow the common practice in \yay{the} computer vision domain~\cite{dosovitskiy2020image} by} \yiz{splitting} the item image into patches, \yiz{flattening} these patches, and linearly \yiz{projecting} them to form patch embeddings. 
Similarly, we obtain the word embeddings from \yay{an} item's \yiz{description}~\cite{radford2019language}.}
% This operation similar to the formation of word embeddings from the text of the item.
Our UGT model \yi{then} uses the patch embeddings from the image and \iadh{the} word embeddings as inputs. 
% Our UGT model \yi{then} uses the patch tokens from the image and \iadh{the} word tokens from the text as inputs. 
Specifically, we \yiz{use} a shared multi-head self-attention component to extract \iadh{the} items' respective \yi{modality} embeddings.
\yay{After} the shared self-attention component,
\yiz{we employ} two 
% \inote{why two, is that because we have two modalities?} 
\yay{specialised} feed-forward networks, \yiz{known as} modality experts, \yiz{corresponding to the two modalities available \yay{in our datasets} -- visual and textual\footnote{Although our UGT model primarily focuses on visual and textual modalities, we \iadh{note that} it can generalise to other modalities \io{when they are present in the datasets.}}.} 
\yiz{These feed-forward networks} \yay{produce} the \yay{corresponding} multi-modal item embeddings.
% And there are two feed-forward networks (i.e., modality experts) used for respective modalities after the shared self-attention module. 
% We route each input token to the experts depending on its modality.
\yi{In our UGT model}, the shared self-attention component in the multi-way transformer \yiz{is designed to align} different modalities, while \iadh{the} modality experts
% \io{the use of} \iadh{the} modality experts \yi{helps} the model 
\yiz{are used} to capture \yiz{information unique to each modality}. 
% ~\cite{li2022blip}
\yay{Consequently, the} multi-way
% \inote{why it is called multi-way, what are these ways?} 
transformer \yay{acts} as a unified feature extractor, \yi{returning} the extracted \io{item's} visual and textual \io{embeddings} \yi{before their combination} with \iadh{the} \yiz{user/item} ID embeddings in a later \yiz{new} unified GNN component. 
\yi{We design this integration between the extraction and fusion \io{components} \io{to address} the isolated extraction problem in exiting multi-modal recommender systems.}
As illustrated in Figure~\ref{fig:arc}, once the extracted multi-modal item features \iadh{have been obtained}, \yay{our model's unified GNN component} jointly \yiz{combines} these \yay{features with} their ID embeddings to \yay{produce} user/item embeddings based on the user-item interaction graph.
The unified GNN is \iadh{needed in order} to obtain a smoother \yiz{user/item} embedding 
% \inote{embedding for what? items, users? both?} 
by aggregating \iadh{the} multi-modal features from \iadh{the} neighbours.
\yiz{This aggregation includes both \iadh{the} user-item interactions and \io{the} semantic information, which is derived from the multi-modal features processed by the multi-way transformer,}
% thereby incorporating both \iadh{the} user-item interactions and \io{the} semantic information \inote{this semantic information from what?} 
into the final user/item representations.
\yi{As \yay{shown} in Figure~\ref{fig:arc}}, within the unified GNN, we incorporate an 
% \yiz{inter-modality} \inote{i do not know what an inter-modality means, besides it is not even shown in fig 2} 
attentive-fusion component to fuse 
% \inote{why further fuse, do you do two fusions then? why further, further to what other fusion?} 
the \ya{jointly-learned} and aggregated embeddings from different modalities, \yi{in order} to enhance the multi-modal user/item representations. 
% in the context of multi-modal recommendation. 
% \yi{As such, the enhanced joint multi-modal features \io{aim} to address the \zx{isolated modality modelling \ya{of}} each modality in multi-modal recommendation. \inote{this sentence is comical; DELETE its just stopwords}} 
\yay{The improved joint multi-modal features in our \yiz{UGT} model are designed to overcome the limitations of modelling each modality separately \yiz{(Isolation 2)} in multi-modal recommendation.} \looseness -1 
% Internally, we also leverage an attentive-fusion component to further fuse the joint-learned/aligned and aggregated embeddings, which aims to facilitate user/item representation learning in multi-modal recommendation.
% Furthermore, within the framework, we employ an attentive-fusion component to additionally merge the jointly learned/aligned and aggregated embeddings, intending to enhance user/item representation learning in the context of multi-modal recommendation.

% As the multi-modal features are jointly learned through this module, the later GNN module is expected to input aligned modal-specific embeddings for further aggregation with their neighbours.
% The unified GNN jointly aggregates the user/item node with their corresponding neighbours with multi-modal item features. 
% The necessity of the unified GNN is to obtain a smoother embedding by aggregating multi-modal features from neighbours, thus incorporating both interaction info and semantic information into the user/item representations.

\vspace{-1mm}
% \pageenlarge{3}
\subsection{Unified Multi-modal Encoding}\label{sec:mmencoding}
As discussed in Section~\ref{sec:mmrec}, we \yiz{propose} a multi-way transformer paired with a fusion component to \iadh{alleviate} the isolated extraction problem \yy{(Isolation 1)} in \yi{existing} multi-modal recommender systems. Specifically, we \ya{use this} multi-way transformer to encode the raw \ya{images and} \yiz{textual descriptions} of items, \yy{replacing} 
% \inote{substituing WHAT with WHAT?} 
the feed-forward network \yiz{with modality experts in} a standard Transformer~\cite{vaswani2017attention}. 
\zx{Using the modality experts allows \ya{to capture} modality-specific information, with each modality expert specifically handling different item modalities.}
% current multi-modal recommender systems depend on pre-extracted multi-modal features from pre-trained extractors, possibly incorporating irrelevant information that could harm the correlation modelling of items in a multi-modal recomm
% Each \zx{modality expert} is expected to capture modality-specific information by selecting a different modality expert. 
\yi{As described in Section\yiz{~\ref{sec:arc}}}, the \yiz{use} of Multi-Head Self-Attention (MHSA), shared across modalities, \yiz{allows} the alignment of \iadh{the} visual and textual features. 
Hence, \yiz{for a given item}, the 
% obtained multi-modal item \inote{remove if for a given item is used} 
embedding is defined as follows:
\begin{equation}\label{eqn:mwtrans}
h_{i}^{(l_{t})} = \text{LN}\left(h_{i}^{(l_{t}-1)} + \text{MHSA}(h_{i}^{(l_{t}-1)}, h_{i}^{(l_{t}-1)}, m_{i}) + \text{FFN}(h_{i}^{(l_{t}-1)}, m_{i})\right)
\end{equation}
where $h_{i}^{(l_{t})}$ is the hidden state of the $i$-th item in the $l_{t}$-th layer, $\text{LN}$ and $\text{MHSA}$ are the layer normalisation operation and the multi-head self-attention mechanism\yiz{~\cite{vaswani2017attention}}, 
% \inote{are there citations for these LN and MHSA?}, 
respectively, \iadh{and} $m_{i}$ is the corresponding multi-modal embedding of the item. $\text{FFN}$ is the feed-forward network.
\io{Hence}, \yi{the multi-way transformer} \yiz{choose\yy{s}} \yay{the right} expert among multiple modality experts to \yay{handle} the input \yiz{based on} \yi{both} the 
% modality of the input vectors 
\yiz{modality's type and}
% and the index of the transformer layer 
\yiz{\yay{the} transformer layer's position}.
% \inote{In our used datasets? Because we use two modalities? .... LINK}
% There \yi{is} \yi{a} vision expert (Visual FFN)
% % \inote{HOW CAN THIS BE ENGLISH< AND HOW YOU DINT SEE THIS YOURSELF} 
% and \yi{a} language expert (Textual FFN), respectively. 
% The choice of a given expert depends on the input modality and the layer within the transformer architecture \inote{did you not already say this two sentences above? if you want to stress this, at least manage the repetition, say Recall that, or as mentioned above ...so at least the reader knows you read your own text!}. 
% The contextualised representations for \yi{the} image-only and text-only inputs are obtained accordingly \inote{unclear, according to WHAT? can you write tghis sentence more simply and avoid passive form?} for subsequent \inote{subsequent is a complex word, avoid over-using} fusions.
\yiz{Note that our UGT model can extract features from any number of modalities, not just two as per our used datasets. The multi-way transformer, combined with the fusion component shown in Figure 2, is designed to unify the extraction and fusion processes, addressing the \yay{problem} of isolated extraction (Isolation 1). }

% \inote{THIS TEXT IS TOO LONG and UNCLEAR; TO REPLACE} It is worthwhile to note that any combination of modalities can be jointly extracted in our UGT model.
% By using the multi-way transformer \yi{as an extractor in our UGT model, \yiz{and integrating it with the subsequent fusion component}, as shown in Figure~\ref{fig:arc}}, we expect to bridge the \yi{gap} between \io{the} extraction and fusion processes in multi-modal recommendation. \yi{This integration \io{aims to address} the isolated extraction problem in the context of multi-modal recommendation.}
% \inote{I suggest to replace with this} \yay{XXXX}

% By integrating the multi-way transformer, we expect to bridge the \yi{gap} between extraction and fusion processes in the context of multi-modal recommendation, \yi{thereby resolving the isolated extraction problem in the context of multi-modal recommendation.}
% The integration of the multi-way transformer allows for more effective extraction capabilities while optimising it with the recommendation objectives \inote{which ones, not mentioned up to now} during the training process.
% \todo{link to figure 2, link to problems being tackled or addressed}

% \pageenlarge{4}
\subsection{Unified Multi-modal Fusion}\label{sec:ugnn}
In Section~\ref{sec:gt}, we \yiz{argued for the need for} a unified approach to process the extracted multi-modal \ya{features} for an effective fusion, \yiz{\yay{instead of} handling each modality individually}. \yay{Hence}, we introduce our \yay{new} unified Graph Neural Network (GNN), \yay{which} \yiz{seamlessly} fuses \iadh{the} image and text embeddings in a joint manner, \yiz{closely integrated with the multi-way transformer.}
% , which uniformly fuses \iadh{the} image and text tokens in a joint manner. 
The propagation function of our unified GNN is defined as follows:
% \begin{equation}\label{eqn:ugnn}
% x_i^{(l_{g})}=(1+\epsilon) \cdot x_{i-vt}^{(l_{g}-1)} + \frac{1}{\sqrt{\mathcal{N}(i) \mathcal{N}(j)}}\left(\sum_{j \in \mathcal{N}(i)} x_{j-vt}^{(l_{g}-1)}+\sum_{j \in \mathcal{N}(i)} x_{j-id}^{(l_{g}-1)}\right)
% \end{equation}
\begin{equation}\label{eqn:ugnn}
x_i^{(l_{g})}=(1+\epsilon) \cdot x_{i-vt}^{(l_{g}-1)} + \left(x_{u-vt}^{(l_{g})}+x_{u-id}^{(l_{g})}\right)
\end{equation}
\looseness -1 where $x_i$ denotes the item representations in \iadh{the} $l_{g}$-th graph convolution layer; $x_{i-vt}$ and $x_{u-vt}$ \yi{denote} the joint multi-modal embeddings of \yi{the item} and its neighbours; $x_{u-id}$ \yiz{denotes} the ID embeddings of \yi{the item's} neighbours;  $\epsilon$ is a scaling factor \cm{that} controls the contribution of an item's self-connection with its associated joint multi-modal feature\yi{s}. \cm{\yay{The user}} embeddings \yay{are similarly} calculated.
% self-connection of a user/item with its corresponding joint multi-modal feature.
To obtain the \yiz{user/item} ID embeddings from the neighbours, we \yiz{use} LightGCN~\cite{he2020lightgcn} to propagate \iadh{the} ID embeddings of \iadh{the} users and items over the \yy{user-item} interaction graph. The propagation function of \yiz{the} ID embeddings \yi{is defined as follows:}

\begin{equation}\label{eqn:gnn}
\begin{aligned}
x_{i-id}^{(l_{g})}=\sum_{i\in\mathcal{N}_{u}}\frac{x_{i-id}^{(l_{g}-1)}}{\sqrt{\left|\mathcal{N}_i\right|\left|\mathcal{N}_u\right|}},
\end{aligned}
\end{equation}
% \inote{simplify to remove P}
where $\mathcal{N}_i$ and $\mathcal{N}_u$ denote the set of neighbours for user $u$ and item $i$, respectively, while $\left|\mathcal{N}_u\right|$ and $\left|\mathcal{N}_i\right|$ represent the size of $\mathcal{N}_u$ and $\mathcal{N}_i$.

A key addition to our UGT model is to propose an attentive-fusion method to enhance the joint embeddings based on the extracted multi-modal features, \yiz{as illustrated in Figure~\ref{fig:arc}}.
% \inote{why dont you refer to Fig 2 where and if this is shown?}.
Given the obtained visual and textual embeddings, $h_{i-v}$ and $h_{i-v}$, derived from Equation\yiz{~\eqref{eqn:mwtrans}}, we combine the multi-modal embeddings with an attentive-fusion approach to obtain a unified joint embedding, as follows:
\begin{equation}\label{eqn:attn}
x_{i-vt}^{(l_{g})}=\alpha h_{i-v}^{(l_{g})} \|(1-\alpha) h_{i-t}^{(l_{g})},
\end{equation}
% \begin{equation}
% x_i^{V T}=\alpha h_{vi} \|(1-\alpha) h_{ti},
% \end{equation}
where $\alpha$ is a learned parameter for 
% \yiz{fusing both the visual embedding $h_{i-v}$ and textual embedding $h_{i-t}$}
the attentive-fusion
and $\|$ represents the concatenation operation. As \io{a consequence}, this fusion approach is \io{designed} to uniformly enhance the resulting user/item embeddings
% from each modality 
\yiz{by weighting the importance of each modality,}
% in Equation~\eqref{eqn:ugnn} \inote{unclear, is Eq (2) about single modalities? it was not clear in text}, 
\io{in contrast to} the simple concatenation operation used in previous  works~\cite{yi2022multi,wei2023lightgt}. 
% \inote{I do not get it, you are also using a concatenation, what different between your concatenation and that of other related work?}. 
%\yi{Once the enhanced joint multi-modal features \cm{have been obtained}, our UGT model is expected to \zx{resolve} the \zx{isolated modality modelling process} in multi-modal recommendation.}
%\ya{Our UGT model uses the enhanced joint multi-modal features to address the problem of isolated modality modelling in multi-modal recommendation.}
\yay{Our UGT model uses the enhanced joint multi-modal features to
address the problem of  isolated modality encoding (Isolation 2) in multi-modal
recommendation.}

% thereby improving the multi-modal recommendation performance in a unified manner. 

% \todo{link to figure 2 and to problems being tackled}

% \subsection{Model Prediction}

% \pageenlarge{4}
\subsection{Model Optimisation}\label{sec:obj}
To ensure the effective extraction and multi-modal fusion \yiz{of the visual and textual modalities}, our UGT model \yay{employs} \iadh{the} Image-Text Contrastive (ITC) loss to \yay{train} the model \yay{in} a joint embedding space where the similarity between an item's image and its corresponding text description is maximised, while \yay{minimising the similarity between mismatched image-text pairs~\cite{li2022blip}}.
%the similarity between mismatched image-text pairs is minimised.
The ITC loss is defined as follows:
\begin{equation}
\label{eqn:itc}
\mathcal{L_\text{ITC}} = -\frac{1}{N}\sum_{z=1}^{N}\log\frac{p_{v2t}(z)}{\sum_{z^{\prime}\neq z}^{N}p_{v2t}(z^{\prime})} - \frac{1}{N}\sum_{z=1}^{N}\log\frac{p_{t2v}(z)}{\sum_{z^{\prime}\neq z}^{N}p_{t2v}(z^{\prime})}
\end{equation}
where $N$ is the batch size, while $p_{v2t}(z)$ and $p_{t2v}(z)$ are the softmax-normalised image-to-text and text-to-image similarities of the $z$-th pair, respectively. 
% Additionally, $z$ denotes the negative pair in the context.
As a result, this ITC loss simultaneously optimises our UGT model during training
% , encouraging the alignment \inote{ditto as before} of \iadh{the} learned embeddings from multiple modalities 
so as to extract and fuse more effective multi-modal features from raw data.
% a joint embedding space is learned using a contrastive loss, where the similarity between an image and its corresponding text encourages the encoder to generate more aligned embeddings.

Following~\cite{yi2022multi,LATTICE21}, we also implement a multi-task training strategy \yiz{in our UGT model} that simultaneously optimises the commonly used pairwise ranking loss \yiz{in recommendation}, specifically the Bayesian Personalised Ranking (BPR)~\cite{rendle2009bpr} \yy{loss}, denoted as $\mathcal{L}_{BPR}$, along with the ITC loss $\mathcal{L}_{ITC}$:
\begin{equation}\label{eqn:multi}
\begin{aligned} 
    \mathcal{L}&=\mathcal{L}_{BPR}+\lambda_{c} \mathcal{L}_{ITC}+\lambda\|\Theta\|_2,\\
    \text{where} \quad \mathcal{L}_{BPR} &= \sum_{(u,i,j)\in D_{s}}\text{log}\sigma ({y}_{ui} - {x_{u}}^{\top}{x_{i}})
\end{aligned}
\end{equation}
\looseness-1 where $\lambda_{c}$ and $\lambda$ are hyper-parameters that control the strengths of the ITC loss and the $L_{2}$ regularisation, respectively, $\Theta$ is the set of model parameters; \yiz{$\boldsymbol{x_{u}}$} is the user embedding, \yiz{$\boldsymbol{x_{i}}$} denotes the positive item embedding and ${y}_{ui}$ is {the} ground truth value, 
% \inote{Big problem: ei eu etc do not appear in the equation, there is a mismatch between text and equation as noted by reviewer}, 
$D_{s}\ =\{(u,i,j)|(u,i)\in R^{+},(u,j)\in R^{-}\}$ is the set of the training data, $R^{+}$ indicates the observed interactions and $R^{-}$ indicates the unobserved interactions, \iadh{while} $\sigma(\cdot)$ is the sigmoid function. 
\yi{\io{Hence}, we \cm{optimise both} the multi-way transformer and the unified GNN \cm{using the} \yi{loss functions} above.
These loss functions prevent the incorporation of non-relevant information that could be harmful to the learned user/item embeddings, further \io{supporting} our UGT model \io{in addressing} the isolated extraction problem \yiz{(Isolation 1)} in a unified manner.}
\yiz{By applying Equation~\eqref{eqn:mwtrans}, we extract multi-modal item embeddings from each modality using the multi-way transformer.}
\yiz{Next, we fuse \yay{the} extracted visual and textual item embeddings using the attentive fusion as detailed in Equation~\eqref{eqn:attn}.}
\yiz{We then integrate the resulting joint item embeddings with the user/item ID embeddings and \yay{the} user-item interactions within our unified GNN, as detailed in Equation~\eqref{eqn:ugnn}, \yay{so as} to obtain the final user/item embeddings.}
% By propagating \io{the} user-item interactions and using the joint multi-modal features, \io{as described} from Equation~\eqref{eqn:attn} through Equation~\eqref{eqn:ugnn} \inote{unclear, reformulate, we do not describe sthg or an equation from X ro Y, it is not English}, we obtain the final user/item embeddings, $x_u/x_i$, which encapsulate multi-modal information from each layer of the multi-way transformer and \io{the} unified GNN. 
% Hence, \yi{after training our UGT \yi{model} by minimising the loss function according to Equation~\eqref{eqn:multi}, 
% we estimate the relevant score \yi{between} a user and \io{the} items \yi{using} Equation~\eqref{eqn:score}}.
% Hence, we can estimate the relevant score \yi{between} a user and items \yi{through} Equation~\eqref{eqn:score} after training our UGT by minimising \yi{loss function} in Equation~\eqref{eqn:multi}. 
\yy{We use} Equation~\eqref{eqn:multi} to optimise both the multi-way transformer and the unified GNN \yy{within our UGT model}, facilitating a unified approach to \io{address} the isolated extraction problem \yiz{(Isolation~1)}.
% \inote{it seems you only solve Isolation X but what about the other one? we need to see two problems solved}.
\io{In addition}, by incorporating an attentive-fusion component, the unified GNN uniformly fuse\yy{s} \io{the} multi-modal features, thereby tackling the \ya{problem of} \zx{isolated modality encoding} \yiz{(Isolation~2)} in multi-modal recommendation.

% \todo[io]{There is a big missing issue; it is not clear if you are focussing on items, or if the same thing is done for usrs, only in one section you mention something about the user being calculated in the same way as items, you need to check if the model deals both with items and users}
% Both the multi-way transformer and unified GNN are optimised through the objective in Equation~\eqref{eqn:multi}. This jointly optimisation on both components can resolve the the isolated extraction problem in a unified manner.
% The unified GNN along with a attentive fusion component can uniformly fuse the multi-modal features, thereby addressing the  independent multi-stream processing problem in multi-modal recommendation.

% Through propagating the user-item interactions and obtained joint multi-modal features (Equation~\eqref{eqn:attn}) with Equation~\eqref{eqn:ugnn}, we obtain the final user/item embeddings $x_u/x_i$ contains multi-modal information from each layer of the multi-way transformer and unified GNN. 
% 1. vlmo  Equation~\eqref{eqn:multi}
% 2. extract -> joint Equation~\eqref{eqn:attn}
% 3. user-item interactions and joint multi-modal features Equation~\eqref{eqn:ugnn}
% 4. score Equation~\eqref{eqn:score} and optim Equation~\eqref{eqn:multi}
% \todo{no link to fig 2, no clear articulation of which problem this hekps to address out of problem 1 and problem 2 in abstract/introduction}

\section{Experiments}\label{sec:exp}
\yay{We} conduct experiments to validate the effectiveness of our UGT \yi{model} on three public datasets, in comparison to \yiz{nine} existing state-of-the-art multi-modal recommendation models.
To examine \yay{and analyse} the effectiveness of our UGT model, we conduct experiments to answer the following \yi{four} research questions:\\
% \begin{itemize}
\noindent  \textbf{RQ1}: How does our proposed UGT model perform compared with \iadh{existing} multi-modal recommendation models?\\
%\noindent  \textbf{RQ2}: \yi{How does each component and the loss of UGT affect the performance of the model?}\\
\noindent \textbf{RQ2}: How do the \yi{two} components \iadh{of} UGT, \yi{namely \io{the} unified GNN and \io{the} multi-way transformer}  
% \inote{what about losses LTC vs LTC + BPR?}
\io{as well as the \yy{UGT's} loss function} 
affect the performance of \yiz{the} model?\\
\noindent  \textbf{RQ3}: How do different parameters (i.e. $\epsilon$, $\lambda_{c}$) in our UGT model affect its performance?\\ 
% \inote{this suggests ITC + BPR is the actual loss function of UGT? this was never clearly said in Sec 3}\\ 
\noindent  \textbf{RQ4}: \yiz{Does the UGT model exihibit a better alignment of modalities compared to \yay{the strongest} baseline \yy{FREEDOM}?\\}

\vspace{-4mm}
\subsection{Experimental Settings}\label{sec:exp}
\begin{table}[tb]
\renewcommand{\arraystretch}{1.2}
\centering
\caption{Statistics of the used Amazon datasets.}\label{tab:stat_all}
% \vspace{-2mm}
\resizebox{0.46\textwidth}{!}{
\begin{tabular}{l|ccc}
     \hline
     & Sports & Clothing & Baby\\
     \hhline{====}
     \#Users &  35,598 &  39,387 &  19,445 \\
     \#Items &  18,287 & 22,499 & 7,037\\
     \#Interactions &295,366 & 271,001 &  160,522\\
     \hline
     Density &0.00045 &0.00030 & 0.00012\\   
     \hline
    CNN/UGT Visual Dimension & 4096/768 & 4096/768  & 4096/768\\
    Sentence-Transformer/UGT Textual Dimension & 384/768 & 384/768 & 384/768\\
    \hline
\end{tabular}}
\vspace{-4mm}
\end{table}

% \pageenlarge{3}
\subsubsection{Datasets}
In order to evaluate the effectiveness of our UGT \yi{model} in the \yi{top-$k$ multi-modal} recommendation task, we conduct experiments on three \yiz{commonly used} Amazon Review datasets\footnote{\url{https://jmcauley.ucsd.edu/data/amazon/}}, namely Sports and Outdoors (Sports for short), Clothing, Shoes and Jewelry (Clothing for short), and Baby.
\ya{Table~\ref{tab:stat_all} shows the} statistics of the used datasets.
% \inote{note that you are using datasets from the same source, so its an easy criticism to have fom reviewers, i thought we want diverse datasets? but you seem to at least justify the choice below}. 
\zx{We \ya{choose} these datasets due to their comprehensive coverage of user-item interactions and \ya{their} abundant multi-modal data, including \ya{the} items' \ya{image} URLs and 
%\ya{the} detailed \ya{items'} 
\yay{textual} descriptions.}
These datasets are \ya{also} unique in providing extensive raw data, unlike other multi-modal recommendation datasets \ya{such as} TikTok\footnote{\url{http://ai-lab-challenge.bytedance.com/}} and Kwai\footnote{https://www.kuaishou.com/activity/uimc}, which do not provide raw data in \yi{relation} \ya{to the} various modalities.
\zx{\ya{In addition}, these datasets have been widely evaluated by several existing state-of-the-art 
% (SOTA) \inote{do you use the SOTA acronym after?} 
baselines, such as~\cite{LATTICE21,zhou2023bootstrap,wei2023lightgt}, 
% \inote{you cite 3 but you use 9? so maybe e.g. ...}, 
further confirming their suitability for our analysis.}
% These datasets include user-item historical interactions and abundant multi-modal data sources such as item image URLs and descriptions, making them suitable for evaluating multi-modal recommendation performance. 
% Other \iadh{existing} multi-modal recommendation datasets such as TikTok\footnote{\url{http://ai-lab-challenge.bytedance.com/}} and Kwai\footnote{https://www.kuaishou.com/activity/uimc} do not provide raw data in respective \yi{relation} of various modalities. 
% \iadh{Different to} existing approaches~\cite{he2016vbpr, wei2019mmgcn, yi2022multi, LATTICE21} that use pre-trained extractors \yi{to extract 4096-dimensional visual features of items~\cite{ni2019justifying}}, we retrieve raw images from the item URLs and encode them into 768 dimensions using our UGT model, 
% \yi{a dimension choice inspired by the original design of the multi-way transformer~\cite{bao2022vlmo}.}
% {For the texts in the datasets}, we concatenate the title, description, brand, and categorical information of items as the textual feature\yi{s}
% and encode it into 768 dimensions with our UGT model. 
% Contrasting with existing approaches that use \zixuan{Sentence-Transformer} to extract 384-dimensional textual embeddings~\cite{LATTICE21}
% Following the dataset processing protocol, widely used in previous works~\cite{LATTICE21, zhou2023comprehensive}, we transform the ratings into binary values of 0 or 1, indicating whether the user has rated the item. 
% In line with previous works~\cite{LATTICE21}, we filter out users and items with more than 5 interactions in a given dataset. 

\subsubsection{Evaluation Protocols}
Following the evaluation setting in ~\cite{LATTICE21, zhou2023comprehensive}, we randomly split the datasets into training, validation, and testing sets using an 8:1:1 ratio.
Both our UGT model and the used baselines have their hyper-parameters optimised using a grid search on the validation set, and 
refined by the Adam~\cite{kingma2014adam} optimiser.
For the other hyper-parameters unique to our UGT model \yiz{($\epsilon$ and $\lambda_{c}$)},
% \inote{name them between parentheses},
we tune \iadh{the parameters as follows:} $\epsilon \in \left \{ 0, 0.1,0.2,...,1.0 \right \}$ \yiz{and} $\lambda_{c} \in \left \{ 0, 0.1,0.2,...,1.0 \right \}$.
% and $\tau \in \left \{ 0, 0.1,0.2,...,1.0,2.0,...10.0 \right \}$ \inote{right, these are not the parameters mentioned in RQ3, where you mention two and a different lambda, confusing}.
We use two commonly used evaluation metrics, namely Recall@K and NDCG@K, to examine the top-$K$ recommendation \iadh{performance}.
\yay{We set} K to \yi{10} and 20~\cite{LATTICE21}, and report the average performance achieved for all users in the testing set. \yay{We also} examine the statistical significance of UGT's improved performance over the used baselines using the paired t-test with the Holm-Bonferroni correction for $p<0.05$.
We apply an early-stopping strategy that terminates the training \yiz{of both UGT and \yay{the} baseline models} if no decrease in validation loss is observed over 50 epochs.
% To perform negative sampling for each user, we identify items that have no prior interactions with the user from the history of observed user-item interactions.
% We report the average performance and examine the statistical significance of UGT's improved performance over the used baselines using the paired t-test with the Holm-Bonferroni correction for $p<0.05$.
% Following the settings in ~\cite{wei2019mmgcn, zhou2023comprehensive}, we use an all-rank item evaluation strategy is used to measure the used metrics.
All used baselines 
% (VBPR\footnote{\url{https://github.com/DevilEEE/VBPR}}, MMGCN\footnote{\url{https://github.com/weiyinwei/MMGCN}}, MMGCL\footnote{\url{https://github.com/zxy-ml84/MMGCL}}, SLMRec\footnote{\url{https://github.com/zltao/SLMRec}}, LATTICE\footnote{\url{https://github.com/CRIPAC-DIG/LATTICE}}) 
and our UGT model are implemented with PyTorch and \yay{were ran} on a GPU A6000 with 48GB memory.
% Our \iadh{source} code, \yiz{processing scripts of raw data} and \yay{the} related documentation are publicly available at: \url{https://anonymous.4open.science/r/UGT-05CE/}.

% \pageenlarge{2}
% \vspace{-2mm}
\subsubsection{Baselines}\label{sec:baseline}
% \begin{table}[tb] 
% \renewcommand{\arraystretch}{1.2}
% % \begin{center}
%     \caption{Summary of \io{the} {compared} approaches.}\label{tab:persp}
% \begin{adjustbox}{width=\linewidth}
% \begin{tabular}{l|c c c c c c c c}
% \cline{1-9}
% Method \ & VBPR & MMGCN & MMGCL & SLMRec & LATTICE & PGMT & LightGT  & UGT\\
% \hline
% Non-GNN-based & \checkmark  & $\times$ & $\times$ & $\times$  & $\times$ & $\times$ & $\times$ & $\times$\\ 
% GNN-based & $\times$  & \checkmark & \checkmark & \checkmark  & \checkmark & \checkmark & \checkmark & \checkmark\\ 
% Graph Transformer & $\times$  & $\times$  &  $\times$ & $\times$ & $\times$ & \checkmark  & \checkmark & \checkmark\\
% \hline
% \end{tabular}
% \end{adjustbox}
% % \vspace{-4mm}
% % \end{center}
% \end{table}
To examine the effectiveness of our UGT model, we perform a comparative analysis between \yi{the baselines}, \iadh{which use} isolated extraction methods and \zx{isolated modality fusion methods}, and our proposed unified UGT model.
%To ensure a fair comparison \iadh{in terms of} model parameters and recommendation effectiveness, \yay{in \zx{the used} baselines}, 
%we \yay{adopt} the widely-used CNN and Sentence-Transformer as \iadh{the} \yi{modality-specific} extractors~\cite{zhou2023comprehensive}. \yay{These extractors} extract \zx{item} \yi{features} from \yi{both the} raw images and \io{the} texts \iadh{in} the used datasets.
\yay{To ensure a fair comparison, we employ the widely-used CNN and Sentence-Transformer as modality-specific extractors in our baselines~\cite{zhou2023comprehensive}. These extractors are \yay{used} to derive item features from the raw images and \yy{descriptions} within the used datasets}.
%Indeed, \yay{in doing so}, both the CNN \& Sentence-Transformer \yay{on one hand and} our multi-way transformer \yay{on the other hand} have a comparable parameter count (170 vs.\ 167 million), ensuring a fair comparison.
\yay{In doing so, we ensure that both the CNN and Sentence-Transformer, as well as our multi-way transformer, have a similar number of parameters (170 million vs. 167 million), ensuring a fair comparison.}
\yay{For our experiments}, we select \yiz{nine} \iadh{existing} state-of-the-art multi-modal recommendation models, namely: GNN-based, \yi{Non-GNN-based} and Graph Transformer models,
% Table~\ref{tab:persp} summarises the baselines across different aspects, 
\yi{as detailed below}:
\yiz{\textbf{(1) \textit{GNN-based \& \yi{{Non}-GNN-based} Models:}}} 
\textbf{VBPR~\cite{he2016vbpr}}, \yiz{a} Non-GNN-based recommendation model, uses VGG net~\cite{gemmeke2017audio} to extract visual features \yiz{and} exclusively integrates these features \yiz{with} \yi{user/item} IDs into matrix factorisation to \yiz{allows} multi-modal recommendation; \textbf{MMGCN~\cite{wei2019mmgcn}} \yiz{uses} graph convolutional networks (GCN) to learn modality-specific features from individual graphs using pre-extracted features. \yay{Then, it} combines these features through simple concatenation to generate final user and item representations;
\textbf{MMGCL~\cite{yi2022multi}} \yiz{also uses} pre-extracted features and
introduces modality edge dropout and modality masking augmentations to the concatenated multi-modal user/item embeddings, enhancing multi-modal representation learning through a self-supervised learning paradigm; 
\textbf{SLMRec~\cite{tao2022self}}, similar to MMGCL, \yiz{is a} GNN-based model that leverages self-supervised learning with pre-extracted features to generate supervised signals by contrasting fine and coarse-item embeddings across each modality;  
\textbf{LATTICE~\cite{LATTICE21}}, \yiz{unlike the aforementioned} GNN-based models, \yiz{builds} item-item graphs from pre-extracted multi-modal features and performs individual graph convolutional operations on both these \yiz{graphs} and the user-item interaction graphs to \yay{produce} user and item representations;
\yiz{\textbf{BM3~\cite{zhou2023bootstrap}} 
% \inote{should BM3 be covered in Sec 2?} 
bootstraps latent user/item representations in contrastive learning by reconstructing the user-item interaction graph to enhance the recommendation performance;}
\yiz{\textbf{FREEDOM~\cite{zhou2023tale}} freezes the structure of LATTICE's item-item graph while denoising the user-item interaction graph by pruning the popular edges.}
\yiz{\textbf{(2) \textit{Graph Transformer Models:}}}
\textbf{PGMT~\cite{liu2021pre1}} \yiz{constructs} an item-item graph to aggregate \yay{the} multi-modal features of items and leverages a transformer network to model the item embeddings along with their contextual neighbours within the obtained graph;
\textbf{LightGT~\cite{wei2023lightgt}} uses a vanilla transformer network to model \iadh{the} users' preferences by considering their interacted items \yiz{along} with individually pre-extracted multi-modal feature\yiz{s} and \yiz{integrates} these preferences with \iadh{the} aggregated \yi{user/item} ID embeddings to predict \iadh{the} user-item interactions. 
\subsection{Performance Comparison (RQ1)}\label{sec:rq1}
\begin{table*}[tb]
\centering
% \vspace{-6mm}
\caption{Experimental results \io{comparing} our UGT model \io{with the} used baselines in multi-modal \io{recommendation}. The best performance of each model is highlighted in bold. $^{*}$ denotes a significant difference compared to the \io{indicated baseline performance} using the paired t-test with the Holm-Bonferroni correction for $p<0.05$. }
% \vspace{-2mm}
\label{tab:comp_base}
\renewcommand{\arraystretch}{1.0}
\begin{adjustbox}{width=\linewidth}
\begin{tabular}{lcccccccccccccc}
\toprule
\multirow{1}{*}{\textbf{Dataset}} & \multicolumn{4}{c}{Sports} & \multicolumn{4}{c}{Clothing} & \multicolumn{4}{c}{Baby}\\ 
\cmidrule(lr){1-1} \cmidrule(lr){2-5} \cmidrule(lr){6-9} \cmidrule(lr){10-13}
Methods & Recall@10 & Recall@20 & NDCG@10 & NDCG@20 & Recall@10 & Recall@20 & NDCG@10 & NDCG@20  & Recall@10 & Recall@20 & NDCG@10 & NDCG@20\\
\midrule
VBPR & ${0.0509}$$^{*}$  & ${0.0771}$$^{*}$  &  ${0.0280}$$^{*}$  & ${0.0349}$$^{*}$   &  ${0.0409}$$^{*}$  & ${0.0611}$$^{*}$  & ${0.0226}$$^{*}$  & ${0.0277}$$^{*}$  & ${0.0479}$$^{*}$  & ${0.0740}$$^{*}$  & ${0.0262}$$^{*}$  & ${0.0329}$$^{*}$ \\
% \midrule
MMGCN & ${0.0290}$$^{*}$   & ${0.0475}$$^{*}$   &  ${0.0154}$$^{*}$  & ${0.0201}$$^{*}$  & ${0.0151}$$^{*}$  &  ${0.0246}$$^{*}$  & {0.0077}$^{*}$  & {0.0100}$^{*}$  & {0.0391}$^{*}$  & {0.0642}$^{*}$  & {0.0201}$^{*}$  & {0.0266}$^{*}$  \\
% \midrule
MMGCL & ${0.0617}$$^{*}$   & ${0.0913}$$^{*}$  & {0.0351}$^{*}$  & {0.0428}$^{*}$  & {0.0410}$^{*}$  & {0.0607}$^{*}$  & {0.0227}$^{*}$  & {0.0277}$^{*}$ & {0.0521}$^{*}$  & {0.0790}$^{*}$  & {0.0283}$^{*}$ & {0.0352}$^{*}$  \\
% \midrule
SLMRec & {0.0605}$^{*}$   & {0.0901}$^{*}$  & ${0.0341}$$^{*}$  & {0.0417}$^{*}$  & {0.0430}$^{*}$  & {0.0623}$^{*}$  & {0.0238}$^{*}$  & {0.0287}$^{*}$  & {0.0527}$^{*}$  & {0.0810}$^{*}$  & {0.0288}$^{*}$  & {0.0361}$^{*}$  \\
% \midrule
LATTICE & {0.0633}$^{*}$   & {0.0944}$^{*}$  & {0.0334}$^{*}$  & {0.0424}$^{*}$  & {0.0484}$^{*}$  & {0.0704}$^{*}$  & {0.0280}$^{*}$  & {0.0336}$^{*}$  & {0.0539}$^{*}$  & {0.0860}$^{*}$  & {0.0291}$^{*}$  & {0.0374}$^{*}$  \\
BM3 & \underline{0.0661}$^{*}$   & \underline{0.0970}$^{*}$  & {0.0350}$^{*}$  & {0.0438}$^{*}$  & {0.0535}$^{*}$  & {0.0797}$^{*}$  & \underline{0.0305}$^{*}$  & \underline{0.0358}$^{*}$  & {0.0542}$^{*}$  & {0.0863}$^{*}$  & {0.0296}$^{*}$  & {0.0380}$^{*}$  \\
FREEDOM & {0.0605}$^{*}$   & {0.0918}$^{*}$  & \underline{0.0352}$^{*}$  & \underline{0.0449}$^{*}$  & \underline{0.0538}$^{*}$  & \underline{0.0809}$^{*}$  & {0.0290}$^{*}$  & {0.0356}$^{*}$  & \underline{0.0551}$^{*}$  & \underline{0.0874}$^{*}$  & \underline{0.0303}$^{*}$  & \underline{0.0385}$^{*}$  \\
\midrule
PMGT & ${0.0524}$$^{*}$  & ${0.0796}$$^{*}$  &  ${0.0303}$$^{*}$  & ${0.0371}$$^{*}$   &  ${0.0403}$$^{*}$  & ${0.0608}$$^{*}$  & ${0.0219}$$^{*}$  & ${0.0264}$$^{*}$  & ${0.0500}$$^{*}$  & ${0.0764}$$^{*}$  & ${0.0271}$$^{*}$  & ${0.0338}$$^{*}$ \\
% \midrule
LightGT & {0.0652}$^{*}$  & {0.0953}$^{*}$  & {0.0347}$^{*}$  & {0.0440}$^{*}$  & {0.0514}$^{*}$  & {0.0755}$^{*}$  & {0.0299}$^{*}$  & {0.0353}$^{*}$  & {0.0544}$^{*}$  & {0.0867}$^{*}$  & {0.0294}$^{*}$  & {0.0381}$^{*}$  \\
\midrule
UGT & \textbf{0.0705}  & \textbf{0.1034} & \textbf{0.0391} & \textbf{0.0477} & \textbf{0.0603} & \textbf{0.0922} & \textbf{0.0330} & \textbf{0.0402} & \textbf{0.0602} & \textbf{0.0930} & \textbf{0.0325} & \textbf{0.0406} \\
\midrule
{\%Improv.} & {6.66}\% & {6.60}\%  & {11.10}\% & {6.24}\%  & {12.08}\% & {13.97}\% & {8.20}\% & {12.29}\%  & {9.26}\% & {6.41}\% & {7.26}\%  & {5.45}\% \\
% UGT & \textbf{0.0682}  & \textbf{0.0972} & \textbf{0.0366} & \textbf{0.0452} & \textbf{0.0552} & \textbf{0.0803} & \textbf{0.0310} & \textbf{0.0373} & \textbf{0.0585} & \textbf{0.0914} & \textbf{0.0313} & \textbf{0.0408} \\
% \midrule
\bottomrule
\end{tabular}
% \vspace{-4mm}
\end{adjustbox} 
\end{table*}

Table~\ref{tab:comp_base} \yay{compares the performance} of our UGT model \yay{to that of} \yiz{nine} multi-modal recommendation baselines.
In the table, the top and second-best results are highlighted in bold and underlined, respectively.
We also evaluate the statistical significance of the difference in performance between our UGT model and \iadh{that of} the used baselines according to the paired t-test with the Holm-Bonferroni correction. 
\iadh{From} the results in Table~\ref{tab:comp_base}, we \yay{observe the following}: \looseness -1
\begin{itemize}[leftmargin=*, noitemsep, topsep=0pt, partopsep=0pt]
    \item \yay{For all three datasets, UGT outperforms} all the baseline models on all metrics by a large margin, and \yay{the differences are statistically significant} in all cases.
    In particular, our UGT model significantly outperforms the strongest baseline, LightGT, \yay{in terms of} Recall@20, showing improvements of \yiz{6.60\%, 13.97\% and 6.41\%} \yay{over the} three \iadh{used} datasets, respectively.
    These significant improvements demonstrate the effectiveness of our graph transformer architecture \iadh{in comparison to} using an \zixuan{independent} multi-modal recommendation model alongside \iadh{the} corresponding pre-trained \yiz{feature} extractors.
    % pre-trained extractors and an isolated multi-modal recommendation model.
    We attribute this improved performance to the \yay{combined} \iadh{effect} of the multi-way transformer and the unified GNN, \yay{along} with the used losses (i.e., BPR loss, ITC loss). \yay{These components} \iadh{enable} \iadh{a} simultaneous optimisation \yay{for} \iadh{extracting} effective multi-modal features, \yay{which are then fused} to enhance \iadh{the} final user/item 
    % \inote{be uniform user-item vs user/item} 
    embeddings.
    % We attribute this improved performance to our multi-way transformer’s ability to extract effective multi-modal features from raw images and texts, and our unified GNN’s capacity for seamless fusion, enabling simultaneous optimisation of both the extracted multi-modal features and their fusion throughout the training process.
    % We attribute the performance to the extraction of raw images and texts with our multi-way transformer and the seamless fusion of our unified GNN so as to simultaneously optimising the extracted multi-modal features and their fusion during the training.

    \item \yay{In Table 2, the comparison of the graph transformer models (PGMT, LightGT, UGT) with the GNN-based models (MMGCN, MMGCL, SLMRec, LATTICE, BM3, FREEDOM), shows that, with the exception of PGMT, the graph transformer models generally exhibit a performance on par with the GNN-based models.}
    %\yay{When comparing} \iadh{the} graph transformer models (PGMT, LightGT, UGT) \iadh{with the} GNN-based models (MMGCN, MMGCL, SLMRec, LATTICE, \yiz{BM3, FREEDOM}) in Table~\ref{tab:comp_base}, we observe that most \iadh{of the} graph transformer models \yiz{show a comparative performance with} \iadh{the} GNN-based models with the exception of PGMT. 
   This result indicates 
   % \inote{does it really?} 
   that the application of the transformer network in \iadh{the} GNN-based models \iadh{is more suitable for} the extraction or distillation of multi-modal features, \iadh{in comparison to} using the transformer for aggregation.
    Similar observations have been made in \yi{other} tasks \yy{such as VQA}, 
    % \inote{e.g.? i.e. name task}, 
    as reported in \cite{he2023multimodal}.
    % This result indicates that the effective application of the transformer network in GNN-based models is to extract or distil multi-modal features instead of using the transformer in aggregation. This is also validated in similar tasks\cite{he2023multimodal}.
    % This suggests that within GNN-based models, the strength of the transformer network lies more in the extraction or distillation of multi-modal features, rather than in aggregation. 
    % This outcome suggests that the judicious application of the transformer network in GNN-based models pertains more effectively to the extraction or distillation of multi-modal features, rather than employing the transformer for aggregation. 
    % This observation is further corroborated by similar tasks 
\end{itemize}
\looseness -1 Hence, in answer to RQ1, we conclude that our UGT model effectively leverage\yy{s} the graph transformer architecture, \yy{consisting of} multi-way transformer paired with a unified GNN, to \iadh{significantly} enhance \io{the} multi-modal recommendation performance \yay{in comparison to 9 strong baselines from the literature}.
%with significant performance improvements.
% Our unified approach presents an end-to-end solution, resolving the isolated extraction process and multi-stream modality processing in multi-modal recommendation.
%\iadh{Indeed, UGT} \io{proposes} an end-to-end solution, \iadh{which bridges} the \zixuan{gap} between the extraction \zixuan{component} and the \zixuan{fusion component}, \iadh{so as} 
%\zixuan{to address the isolated extraction problem in multi-modal recommendation.}
\yay{UGT indeed offers a comprehensive end-to-end solution that effectively and seamlessly combines the extraction and fusion components, thereby \yay{addressing} the issue of isolated \yy{feature} extraction (Isolation 1)
% \inote{please tell me if this isolation 1 2 or both between parentheses (Isolation X or Y or Z and Y)}
in multi-modal \yy{recommender} systems}.

% \pageenlarge{3}
% \zixuan{to learn more effective extracted multi-modal features and subsequent user/item embeddings for the multi-modal recommendation task.}
% while also jointly fusing the extracted multi-modal features in a unified approach. 
% which provides an end-to-end solution to overcome the isolation between extraction and recommendation and jointly and seamlessly fuse the extracted feature in a unified manner.

% \pageenlarge{2}
\vspace{-2mm}
\subsection{Ablation Study (RQ2)}\label{sec:rq2}
\begin{table*}[tb]
\centering
% \vspace{-6mm}
\caption{Ablation study on \yay{the} key components of UGT. $^{*}$ indicates the significant differences according to the paired-t test.}
\vspace{-2mm}
\label{tab:ablation}
\renewcommand{\arraystretch}{1.0}
\begin{adjustbox}{width=\linewidth}
\begin{tabular}{lcccccccccccccc}
\toprule
\multirow{1}{*}{\textbf{Dataset}} & \multicolumn{4}{c}{Sports} & \multicolumn{4}{c}{Clothing} & \multicolumn{4}{c}{Baby}\\ 
\cmidrule(lr){1-1} \cmidrule(lr){2-5} \cmidrule(lr){6-9} \cmidrule(lr){10-13}
Methods & Recall@10 & Recall@20 & NDCG@10 & NDCG@20 & Recall@10 & Recall@20 & NDCG@10 & NDCG@20  & Recall@10 & Recall@20 & NDCG@10 & NDCG@20\\
\midrule
UGT w/o Attn-Fuse & {0.0675}$^{*}$  & {0.0986}$^{*}$ & {0.0375}$^{*}$ & {0.0464} & {0.0564}$^{*}$ & {0.0823}$^{*}$ & {0.0310}$^{*}$ & {0.0373}$^{*}$ & {0.0568}$^{*}$ & {0.0894}$^{*}$ & {0.0308}$^{*}$ & {0.0404} \\
UGT w/o UGNN & ${0.0667}^{*}$  & ${0.0980}^{*}$ & {0.0371}$^{*}$ & {0.0452}$^{*}$ & ${0.0540}^{*}$ & ${0.0811}^{*}$ & ${0.0302}^{*}$ & ${0.0356}^{*}$ & ${0.0551}^{*}$ & ${0.0865}^{*}$ & ${0.0298}^{*}$ & ${0.0383}^{*}$ \\
UGT w/o Trans & {0.0628}$^{*}$  & {0.0929}$^{*}$ & {0.0357}$^{*}$ & {0.0440}$^{*}$ & {0.0483}$^{*}$ & {0.0784}$^{*}$ & {0.0269}$^{*}$ & {0.0323}$^{*}$ & {0.0535}$^{*}$ & {0.0834}$^{*}$ & {0.0290}$^{*}$ & {0.0364}$^{*}$ \\
UGT w/o CL & {0.0680}$^{*}$  & {0.0953}$^{*}$ & {0.0381}$^{*}$ & {0.0458}$^{*}$ & {0.0578}$^{*}$ & {0.0896}$^{*}$ & {0.0309}$^{*}$ & {0.0360}$^{*}$ & {0.0543}$^{*}$ & {0.0855}$^{*}$ & {0.0303}$^{*}$ & {0.0377}$^{*}$ \\
\midrule
UGT & \textbf{0.0705}  & \textbf{0.1034} & \textbf{0.0391} & \textbf{0.0477} & \textbf{0.0603} & \textbf{0.0922} & \textbf{0.0330} & \textbf{0.0402} & \textbf{0.0602} & \textbf{0.0930} & \textbf{0.0325} & \textbf{0.0406} \\
\bottomrule
\end{tabular}
% \vspace{-4mm}
\end{adjustbox} 
\end{table*}

\looseness-1 
%\yi{In this section, we evaluate the impact of each of the two components of our \yay{UGT} model, namely the multi-way transformer (\cm{denoted \io{by} Trans in this section}) and the unified GNN (UGNN) \io{component}, \yay{on the recommendation performance}.} 
\yay{In this section, we evaluate how each component of the UGT model, the multi-way transformer (referred to as Trans) and the unified GNN (UGNN), influences the recommendation performance.} Recall that the unified GNN component incorporates an attentive-fusion (Attn-Fuse) as an internal component \yy{(see Fig. 2)}.
% \inote{refer to Fig. 2 e.g. (See Fig. 2)}.
% In this section, we aim to investigate the impact of each of the two components of our model, namely the multi-way transformer (Trans), and the unified GNN (UGNN), with the unified GNN incorporating attentive-fusion (Attn-Fuse) as an internal component.
Specifically, \yay{to evaluate the impact on effectiveness,} we \yay{replace} the multi-way transformer (Trans) and the unified GNN (UGNN) components with a CNN + Sentence Transformer\zixuan{~\cite{zhou2023comprehensive}} extractor and LightGCNs\zixuan{~\cite{he2020lightgcn}} respectively, \yy{\yay{which are} commonly used in the baseline models}. 
% \inote{why these, is that because they are the ones typically used in the baselines?}, 
%to evaluate the \iadh{impact} of these two components on effectiveness.
\iadh{In addition}, we simplify the attentive-fusion (Attn-Fuse) component \yay{by changing its} attentive concatenation \yay{to a straightforward} concatenation of the extracted multi-modal features \yi{so as to assess the effectiveness of the attentive-fusion component}.
\yay{Finally}, we evaluate the effect of contrastive learning (CL) within the UGT's loss function by removing the contrastive loss \yay{(c.f. Equation~\eqref{eqn:multi})}.
Table~\ref{tab:ablation} \yay{presents} the \yay{performance outcomes of UGT's variants} in multi-modal recommendation \yay{across the} three datasets \yay{we used}.
We observe the following:
% \pageenlarge{2}
\begin{itemize}[leftmargin=*, noitemsep, topsep=0pt, partopsep=0pt]
    \item \iadh{Comparing} UGT to its variant \yay{without the attentive concatenation}  (UGT w/o Attn-Fuse), we \iadh{note} 
    %that the removal of \iadh{the} attentive-fusion  \iadh{component} leads to 
    a marked decrease in effectiveness \yay{of UGT} across all \cm{three} datasets. This result indicates the importance of our attentive-fusion component in effectively fusing the multi-modal features into \iadh{the} final user/item embeddings. \looseness -1
    \item \looseness -1 \yay{The ablation of the unified GNN (UGNN) and its replacement with a LightGCN multi-stream processing on each modality~\footnote{\iadh{Indeed, most GNN-based models (MMGCL, SLMRec, LATTICE) apply LightGCNs in multi-stream processing}, leads to a significant decrease of UGT's performance.}.}
    %\cm{Next, we} ablate the unified GNN (UGNN) of our UGT model, \iadh{and \yy{replace}} it with \iadh{a} multi-stream processing using LightGCNs on each modality~\footnote{\iadh{Indeed, most GNN-based models (MMGCL, SLMRec, LATTICE) apply LightGCNs in multi-stream processing}.}. \cm{We}
    %observe that the recommendation performance significantly benefits from the integration of the unified GNN. 
    This \cm{observation} emphasises the essential role of \cm{applying} \ya{a} unified processing \iadh{on the} multi-modal inputs, \iadh{in contrast to} the \yay{traditional method} \yay{that processes each modality input \zx{separately}}.
    % through \yi{multiple independent} methods.
    \item Table~\ref{tab:ablation} \yay{also shows} \iadh{that} \yi{replacing} our multi-way transformer (Trans) \yi{component} \yi{with a CNN + Sentence Transformer extractor} \yay{markedly} degrades the \yy{UGT's} recommendation performance. \iadh{This result} highlights the advantage of \yy{replacing} \iadh{the} pre-trained extractors, \yi{typically used in the existing methods}, with our multi-way transformer, \yi{thereby}
    % \inote{What the latter refers to? something changed, this is vague now} 
    \yi{enabling the} simultaneous optimisation of the extraction process and the generation of \iadh{the} user/item 
    % \inote{hyphen or /} 
    embeddings \zixuan{\yay{for effective} multi-modal recommendation}. \looseness -1
    \item 
    % \inote{You already introduced this} We also introduce a UGT variant, $UGT \, w/o \, CL$, to investigate the impact of the used contrastive loss. 
    \yay{The} UGT model, \yy{which uses both the ITC and BPR losses}, significantly \yy{outperforms} \yy{\yay{its} variant without contrastive learning} (i.e., \zx{UGT}$\, w/o \, CL$). This suggests that the contrastive loss, as introduced in Sec~\ref{sec:obj}, \yy{is beneficial 
    for multi-modal recommendation.} 
    \yy{Indeed, CL,} simultaneously provides large gradients to 
    \yy{optimise both the extraction (multi-way transformer) and fusion (\yay{UGNN}) components in our UGT model, thereby improving the recommendation performance.}
    % \zixuan{enhance} \zixuan{both the extraction of \iadh{the} multi-modal features and the fusion of the final user/item embeddings \inote{can we really say this? is that exactly what CL does?},}
    % % extracted multi-modal features during training, 
    % thereby improving the recommendation performance.
\end{itemize}
% \pageenlarge{2}
\looseness -1 Hence, in answer to RQ2, we conclude that UGT successfully
\ya{uses} \cm{each of its key} components \io{as well as its} \yi{loss function} to provide an effective unified approach for learning effective user/item
% \inote{hyphen or /} 
representations in multi-modal recommendation.
Specifically, our unified GNN (UGNN) component effectively fuses the multi-modal features \zixuan{using} an attentive-fusion (Attn-Fuse) method, thereby addressing the \ya{problem of} \zx{isolated modality encoding} \yiz{(Isolation 2)} \ya{in the existing multi-modal recommenders}.
Furthermore, when we replace the CNN + Sentence Transformer extractor with our multi-way transformer (Trans) component \yy{and pair it with the unified GNN}, we observe an improvement in \yay{recommendation performance}, \yy{thereby addressing the isolated feature extraction problem (Isolation~1).}
% \inote{does this relate to any isolation 1 or 2?}.
% our multi-way transformer (Trans) component enhances the performance of multi-modal recommendation when substituted for the CNN + Sentence Transformer extractor \inote{cannot parse this sentence, substitured means replaced, what is replaced by what? substituted for is unclear to me} .
% benefits the multi-modal recommendation tasks when substituting the CNN + Sentence Transformer extractor with our multi-way transformer component.
\cm{Finally}, the contrastive loss (CL) effectively 
\yy{optimises both the unified GNN (UGNN) and \io{the} multi-way transformer (Trans) \iadh{components}, thereby improving the recommendation performance.}
% enhances both the unified GNN (UGNN) and \io{the} multi-way transformer (Trans) \iadh{components} in our applied \yi{loss functions}, \yy{as detailed in Equation~\eqref{eqn:multi}} \inote{do we really conclude this? can this be made a bit more to the point? we already had one bullet point on CL, so be SHORT and to the point}.

% leverages the unified GNN layers with its internal attentive-fusion component, multi-way transformer and contrastive loss \inote{unclear how this relates to UGT components; you mixing compoents with loss etc; separate into 2 different conclusions as these are separate issues} to provide a unified approach for learning effective user/item representations in multi-modal recommendation.
% In particular, our proposed unified GNN successfully addresses the independent multi-stream processing problem by effectively fusing the multi-modal features \zixuan{using} an attentive-fusion component, \zixuan{thereby leading} to an improved performance.

% \pageenlarge{4}
\subsection{Hyper-parameter Study (RQ3)}\label{sec:rq3}
\begin{figure}[tb]
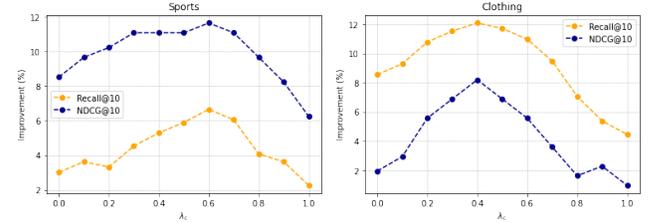

    \begin{subfigure}[t]{0.49\linewidth}
        \includegraphics[trim={0cm 0cm 0cm 0cm},clip,width=1\linewidth]{Sports.pdf}
    \end{subfigure}
    \begin{subfigure}[t]{0.49\linewidth}
        \includegraphics[trim={0cm 0cm 0cm 0cm},clip,width=1\linewidth]{Clothing.pdf}
    \end{subfigure}
    %     \begin{subfigure}[t]{0.33\linewidth}
    %     \includegraphics[width=1\linewidth]{Baby.pdf}
    % \end{subfigure}
    \caption{Performance of our UGT model with respect to different $\lambda_{\yy{c}}$ on the Sports and Clothing datasets.}
\label{fig:hyper}\vspace{-1em}
% \vspace{-5mm}
\end{figure}

\begin{figure}[tb]
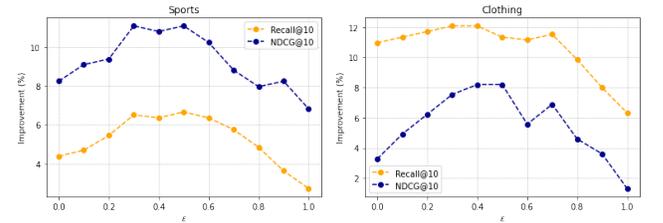

    \begin{subfigure}[t]{0.49\linewidth}
        \includegraphics[trim={0cm 0cm 0cm 0cm},clip,width=1\linewidth]{Sports2.pdf}
    \end{subfigure}
    \begin{subfigure}[t]{0.49\linewidth}
        \includegraphics[trim={0cm 0cm 0cm 0cm},clip,width=1\linewidth]{Clothing2.pdf}
    \end{subfigure}
    %     \begin{subfigure}[t]{0.33\linewidth}
    %     \includegraphics[width=1\linewidth]{Baby.pdf}
    % \end{subfigure}
    \caption{\io{Performance} of our UGT model \io{using} different $\epsilon$ \io{values} on the Sports and Clothing datasets.}
\label{fig:hyper2}\vspace{-1em}
\vspace{-2mm}
\end{figure}

\yay{We now} study the sensitivity of our UGT model to \iadh{the} hyper-parameters. \yay{We focus on Recall@10} for \yay{reporting} the recommendation performance, \yay{since we observe the same trends and} conclusions \yay{with NDCG@10} across \iadh{the used} datasets.
% In this section, to guide the selection of hyper-parameters of our UGT model, we perform a hyper-parameter sensitivity study with regard to the recommendation performance, in terms of Recall@10, since using NDCG@10 also leads to the same conclusions across the used datasets.
For brevity, \yay{Figure~\ref{fig:hyper} presents} the effects of \iadh{the} hyper-parameters \yay{for} the Sports and Clothing datasets, since the results \iadh{on} the Baby dataset lead to similar \zixuan{conclusions}.
% For conciseness, in Figure~\ref{fig:hyper}, we only report the effectiveness of UGT on the Sports and Clothing datasets, since we observe similar conclusions on the Baby dataset.
We primarily analyse two important parameters in our UGT model, \iadh{namely: (i)} the scaling factor $\epsilon$, which controls the influence of an item's intrinsic multi-modal feature\zixuan{s} in the graph convolution operations; and \iadh{(ii)} the contrastive factor $\lambda_{c}$, regulating the strength of the Image-Text Contrastive (ITC) loss during training.
% Specifically, we consider the following two hyper-parameters as they are the most important parameters related to our proposed UGT model, i.e. the scaling factor $\epsilon$ to control the contribution of the item's own multi-modal feature in the graph convolution operations and the contrastive factor $\lambda_{c}$ which control the strength of ITC loss during the training phase.
Figure~\ref{fig:hyper} and Figure~\ref{fig:hyper2} show the performance of UGT with different values of $\epsilon$ and  $\lambda_{c}$, respectively.

% \pageenlarge{3}
\subsubsection{Impact of \iadh{the} contrastive factor $\lambda_{c}$}
The contrastive factor $\lambda_{c}$ indicates the importance of the contrastively learned loss, 
balancing the contribution of \iadh{the} ITC loss within the joint training loss in Equation~\eqref{eqn:multi}.
% between item visual and textual features in the joint training loss. 
\yay{As mentioned in Section~\ref{sec:exp}, we} \iadh{vary} $\lambda_{c}$ within the range \{0.0, 0.1, ..., 0.9, 1.0\} \yy{with a step size of 0.1}, %\iadh{following} our experimental setup in Section~\ref{sec:exp}.
\io{From} Figure\yi{~\ref{fig:hyper}}, 
% \inote{Which figure? 3 or 4?}, 
we observe that the \iadh{best} performance of our UGT model \iadh{occurs} at $\lambda_{c}$ = 0.6, \iadh{and at} $\lambda_{c}$ = 0.4 on the Sports and Clothing datasets, respectively, with a marked performance degradation at higher $\lambda_{c}$ values.
\yi{A high $\lambda_{c}$ \io{value} indicates \io{that} UGT emphasises \io{the} \io{item's} image-text similarity \yy{as measured by the ITC loss}, over the \yy{BPR pairwise ranking loss in the} recommendation task.}
% \inote{CHECK, is that precise you need to be precise}.
% with a marked degradation in performance at higher $\lambda_{c}$ values.
These \iadh{results} \iadh{highlight} the \yay{importance} of carefully \yay{choosing} $\lambda_{c}$ to \yay{tailor} the loss \yi{function} \yay{to} the task, thereby improving the \yay{model's} performance. 
% \inote{why is lambda too high a problem - what does high lambda do intuitively?}
% \inote{You need to say what lambda was learned automatically for getting your RQ1 results, and whether what was leanrned was good enough; in other words, can the parameter be set automatically using adequate training}

\vspace{-2mm}
% \pageenlarge{3}
\subsubsection{Impact of \iadh{the} multi-modal scaling factor $\epsilon$}
To \yay{improve} the multi-modal fusion \yay{within} our unified GNN, we \yay{introduced} a scaling factor, $\epsilon$, \yay{which incorporates} \cm{each} item's own multi-modal features into the graph convolution operations, as \yay{detailed} in Equation~\eqref{eqn:ugnn}. 
To analyse the effect of incorporating an item's own multi-modal \io{features} in the graph convolution operations, we vary $\epsilon$ in the range \{0.0, 0.1, ..., 0.9, 1.0\} with a step size of 0.1,
\yy{as previously described in Section~\ref{sec:exp}.}
% \inote{did you already say this in experimental setup, if so ack that}.
From Figure~\ref{fig:hyper2}, we observe that our UGT model reaches its peak performance on the Sports and Clothing \iadh{datasets} when $\epsilon$ = 0.5 and $\epsilon$ = 0.4, respectively. 
% Figure~\ref{fig:hyper2} shows that the most appropriate value of $\epsilon$ in our UGT model ranges in ${0.4, 0.5}$.
This suggests \iadh{that} the \yay{best} value $\epsilon$ \yay{differs} slightly depending on the recommendation scenario. 
However, the performance drops markedly for a higher value \iadh{of} $\epsilon$.
These results \yay{indicate} that the inclusion of an item's \yay{individual} multi-modal \io{features} effectively enhance\yy{s} the user/item embedding\yy{s}
% \inote{hypthen or /} 
\yay{within a} multi-modal recommendation \yay{system}.
\yi{However, excessively \io{weighting} \io{the} item's own \io{features} \yi{(indicated by a high value of $\epsilon$)} will \yi{decrease} the recommendation performance.}
% adding an item's own multi-modal feature  with the inclusion of an item's multi-modal feature, 
% while excessively incorporating its own feature \yi{into the final embedding} will \yi{decrease} the recommendation performance \inote{reformulate}. \inote{Agasin, you do not say if for answering RQ1, the model learned a reasonable epsilon value}

% \pageenlarge{4}
\yay{Note that for} the \iadh{results} in Table~\ref{tab:comp_base}, UGT learned effective values for $\epsilon$ and $\lambda_{c}$ ($\epsilon$ =  0.5, $\lambda_{c}$ = 0.6 on the \iadh{Sports} dataset and $\epsilon$ = 0.4, $\lambda_{c}$ = 0.4 on the Clothing dataset).
This \yay{shows} that our UGT model \yay{is capable of} automatically \yi{determining} \yay{suitable} values for $\epsilon$ and $\lambda_{c}$ \yay{through a} grid search on the validation data (as per Figures~\ref{fig:hyper} and \ref{fig:hyper2}).
Hence, in answer \yay{to} RQ3, we find that the performance of UGT 
%\yy{remains relatively \inote{overall fairly} stable across these hyper-parameters.}
is \io{overall} \yay{fairly} sensitive to the changes in \iadh{the} $\epsilon$ and $\lambda_{c}$ parameters.
% \inote{can we actually say which parameter the model is more sensitive to? or we cannot really from the results?}. 
\iadh{\io{Indeed, while}} our UGT model \yay{is able to} automatically learn adequate values for different datasets, \iadh{our parameter analysis shows that a} careful selection of these parameters is \yay{in general} needed for \yy{optimal} results.
% However, our UGT model can automatically determine adequate values for different datasets
% \todo{you do not answer explicitely RQ3 like you do in the previous sections}

% \vspace{-2mm}

\begin{figure*}[tb]
    \begin{subfigure}[t]{0.245\linewidth}
        \includegraphics[trim={0cm 0cm 0cm 0cm},clip,width=1\linewidth]{fig_align_extract_sports.pdf}
    \end{subfigure}
    \begin{subfigure}[t]{0.245\linewidth}
        \includegraphics[trim={3.2cm 0.2cm 3cm 0.2cm},clip,width=1\linewidth]{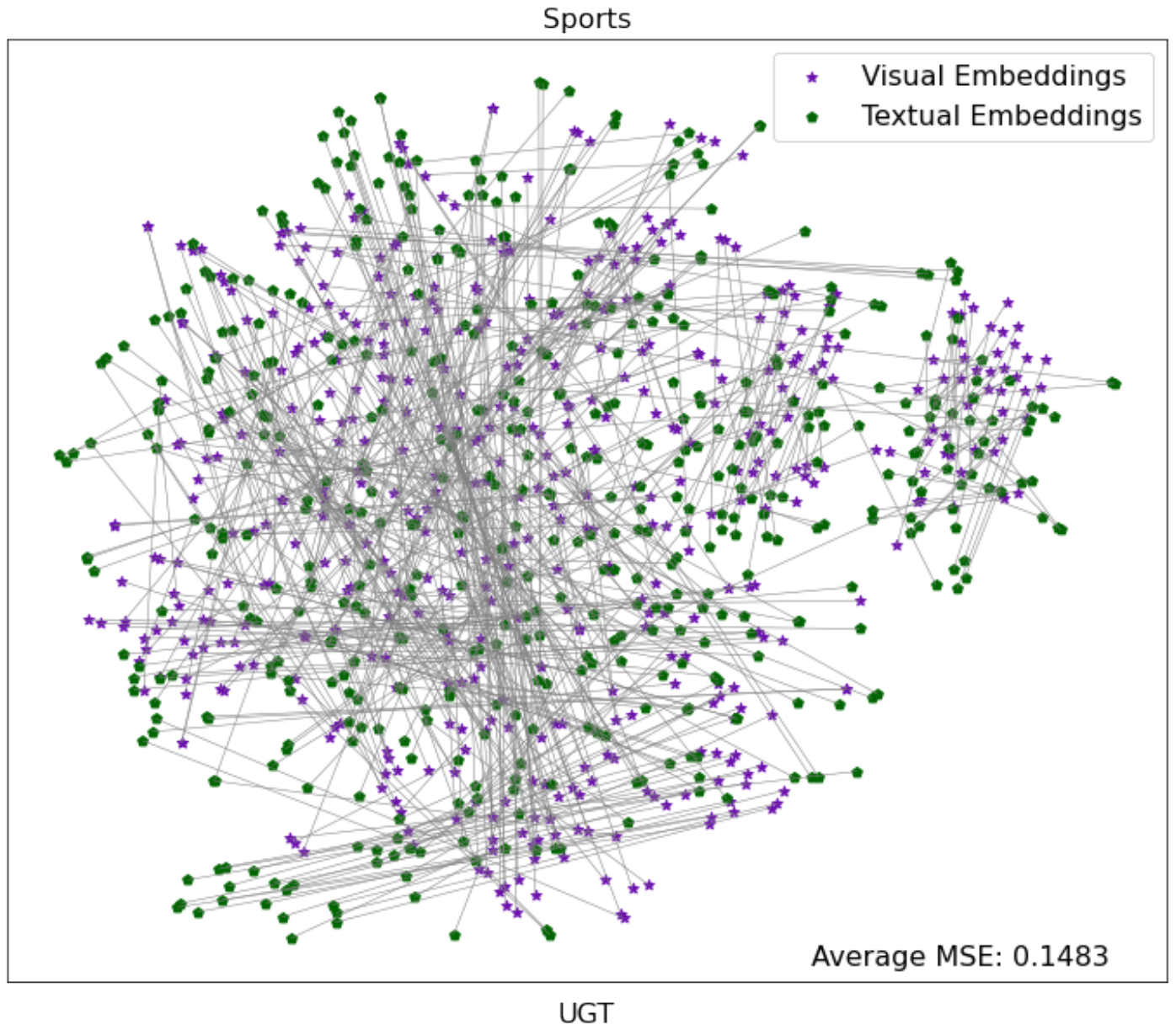}
    \end{subfigure}
    \begin{subfigure}[t]{0.245\linewidth}
        \includegraphics[trim={0cm 0cm 0cm 0cm},clip,width=1\linewidth]{fig_align_extract_clothing.pdf}
    \end{subfigure}
    \begin{subfigure}[t]{0.245\linewidth}
        \includegraphics[trim={3.2cm 0.2cm 3cm 0.2cm},clip,width=1\linewidth]{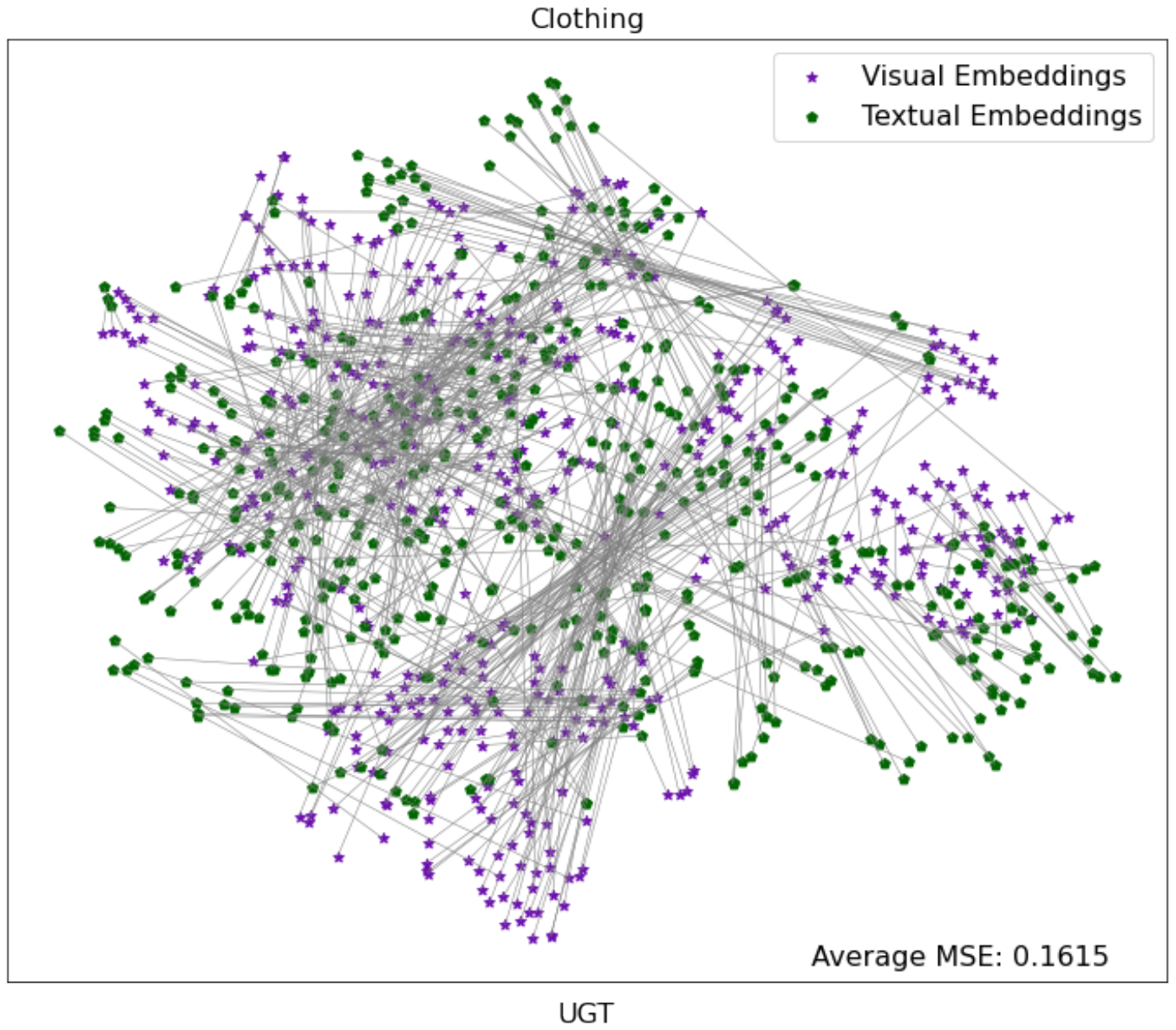}
    \end{subfigure}
    \caption{The t-SNE visualisation of \ya{the} item embeddings on the Sports and Clothing datasets. The star \ya{refers to} the visual embeddings \ya{while} the pentagon represents the text embeddings. The average Mean Squared Error (MSE) value indicates the average distance between \ya{the} visual and textual embeddings.}
\label{fig:tsne}
\vspace{-4mm}
\end{figure*}

% \pageenlarge{3}
% \vspace{-1mm}
\subsection{Modality Alignment of UGT (RQ4)}\label{sec:rq4}
\yy{\yay{We have already shown in Table ~\ref{tab:comp_base} that our UGT model, which addresses both the Isolation~1 and Isolation~2 problems} outperforms the existing baselines}. \yy{In this section, we \yay{analyse whether the improved performance is reflected in the quality of features produced by our UGT model in comparison to those generated by the baselines. In the following, we examine the most competitive baseline, namely FREEDOM. However, we note that we observed similar trends and conclusions with other baselines.}}
\yy{\yay{To assess the quality of the generated features, we measure the average Mean Squared Error (MSE) between the visual and textual embeddings produced by UGT and FREEDOM. \yay{This metric allows} to \yay{assess how well} \yay{the} multi-modal embeddings \yay{align}.}}
% This metric helps us understand how closely these models align multi-modal embeddings.
\yy{We also use t-SNE visualisations~\cite{van2008visualizing} to show the visual and textual embeddings from UGT and FREEDOM across the used datasets.}
Figure~\ref{fig:tsne} shows the \ya{visualisations} of the obtained visual and textual embeddings \ya{on the Sports and Clothing datasets}, with stars and pentagons representing the visual and textual embeddings, respectively.
\yiz{\yay{Due to space limitations}, we omit the results \yy{from} the} Baby dataset since it exhibits the same \ya{trends and conclusions}.
\yy{In Figure~\ref{fig:tsne}, we link \yay{the} two types of embeddings with a light grey line to show their pairwise relationship.}
%\yy{We expect better-quality multi-modal features to exhibit cohesive distributions and lower MSE values, indicating an effective fusion of different modalities into a unified representation.}
\yay{We anticipate that higher-quality multi-modal features will exhibit cohesive distributions and reduced Mean Squared Error (MSE) values, indicating a successful integration of various modalities into a unified representation.}
\yy{Conversely, poorer-quality features \yay{will likely} appear more dispersed and \yay{have} higher MSE values, suggesting an insufficient alignment between \yay{the} modalities.}
% \yy{We expect better-quality multi-modal features to exhibit cohesive distributions, whereas poorer-quality features might appear more dispersed.}
\yy{From Figure~\ref{fig:tsne}, we observe that the visual embeddings of \yay{items in} FREEDOM are densely clustered in the central space of the Sports dataset while their corresponding textual embeddings are more dispersed around this central cluster.}
\yy{In contrast, the UGT model \yiz{produces} visual and textual embeddings that are cohesively distributed, indicating a \yay{tighter} semantic space on the Sports dataset.}
\yy{The markedly lower average MSE value (0.1483) for the UGT model compared to \yay{that of} FREEDOM \yay{10.3732} on the Sports dataset further demonstrates that UGT \yay{better} aligns \ya{the} visual and textual embeddings \yay{than FREEDOM}.}
% \pageenlarge{3}
\yy{We \yay{observe} similar \yay{trends} \yay{in} the Clothing dataset \yay{as shown} in Figure~\ref{fig:tsne}, where \yay{the embeddings of} UGT are cohesively distributed, \yay{in contrast to} FREEDOM's embeddings, \yay{which} are \yay{more widely scattered} with their corresponding pairs.}
\yy{\yay{Indeed,} UGT exhibits a lower average MSE value (0.1615) compared to FREEDOM (8.4460), confirming that UGT more effectively unifies the \yy{multi-modal} embeddings into a \yay{tighter} semantic space.}
\yy{\yay{Overall, these} observations indicate that UGT, by \yay{addressing} the issues of isolated feature extraction and modality encoding through a unified approach, produces higher-quality multi-modal features than FREEDOM.
The \yay{tighter} alignment of \yay{the} visual and textual embeddings \yay{within UGT} \yay{is reflected} in \yay{its} improved performance, highlighting \yy{the importance of better aligning the multi-modal data for an effective} multi-modal recommendation \yay{system}.} \looseness -1

\zx{Hence, in response to RQ4, our UGT model \yy{aligns the multi-modal embeddings} \yy{more cohesively than} the strongest baseline FREEDOM, \yy{resulting in higher-quality features \yay{and a more enhanced recommendation performance}.}}

\noindent \textbf{Supplementary Experiments:}
\yy{In the supplementary material of this paper, we also conduct a \yay{user case} study \yay{using} our UGT model. \yy{We also analyse the contribution of each modality to the performance of the system} for both the UGT and baseline models.}
\yy{We show results in the user case study demonstrating that our UGT model identifies the visual and textual semantics of potential items more effectively than the strongest baseline FREEDOM.}
\yy{\yay{Our study of the contribution of each modality to the performance of the system shows} that our UGT model \yay{more effectively aggregates} the visual and textual features \yay{in comparison to all used baselines}, \yay{thereby obtaining} more effective user/item embeddings.}
\yy{Due to page constraints, we include these details in the \yay{provided} supplementary material.}

\vspace{-2mm}
\section{Conclusions}
\yay{We} proposed the \yay{Unified \yy{m}ulti-modal Graph Transformer (UGT)} model to \zixuan{address} the \zx{problems} of \ya{the} isolated extraction process \yy{(Isolation 1)} and \ya{the} \zx{isolated modality encoding} \yy{(Isolation 2)} in \yay{existing} multi-modal recommendation \yay{models}. \yay{UGT uses a novel} graph transformer architecture \yay{including two new key components}. Specifically, \yay{\yy{t}he UGT architecture} \iadh{addressed} the isolated extraction problem by integrating a multi-way transformer as an extraction component \yy{\yay{before later integrating} it} \yay{to a new} unified GNN, \yay{which serves} as a fusion component.
% \yi{Moreover, we \io{optimised} both of these components together with the same loss functions to further enhance the multi-modal recommendation performance.}
\yay{The new unified GNN allows also to address the \zx{problem of isolated modality encoding}. This component uniformly fuses the extracted multi-modal features.} 
%We also \io{addressed} the \zx{problem of isolated modality encoding} by proposing a unified GNN as the fusion \io{component} \zx{within our UGT architecture} to uniformly fuse the extracted multi-modal features.
\yay{The} obtained multi-modal features \yay{are then combined} using an attentive-fusion method \yay{to further enhance the final user/item embeddings for effective multi-modal recommendation.} %enhancing the effective modality fusion of the final user/item embeddings in multi-modal \iadh{recommendation}.
Our results on three benchmark datasets showed that our UGT model effectively leverages the graph transformer architecture in a unified approach, resulting in a significant performance improvement compared to \yiz{nine} strong baseline models. 
\yi{The \io{performance} improvement reaches up to \yiz{13.97\%} \io{in comparison} to the strongest baseline model (i.e., FREEDOM).}
% \inote{ by up to 22.12\%   ... sadd another sentence saying this 22 per cent is wrt to what? best baseline? strongest baseline?}.
\iadh{In addition}, we conducted an ablation study to \yay{demonstrate} the positive impact of each component of our UGT model \iadh{on the recommendation performance}.
\zx{Finally, we \yy{measured \yay{and visualised} the average MSE distances between} the visual and textual embeddings of items from UGT and FREEDOM. \yay{We found} that UGT produces higher-quality features than FREEDOM, indicating a better}  alignment between the modalities, \yay{and further confirming UGT's recommendation effectiveness}.
% our UGT model effectively aligns different modalities into a unified semantic space compared to \yay{a} \yy{strongest} baseline (i.e., \yy{FREEDOM}).

% \zx{We also visualised the visual and textual embeddings, which illustrated that, in comparison to LightGT,
% % demonstrating through this visualisation that 
% our UGT model can effectively align different modalities into a unified semantic space.}
% \zx{\ya{Finally,}} we \ya{presented} a case study \ya{that illustrates} the effectiveness of our UGT model \iadh{in identifying} useful visual and textual information to \io{a} target user.

% \zx{We also visualised the visual and textual embeddings and performed a case study on the used datasets to show that our UGT model can effectively align different modalities into one shared semantic space, thus identifying useful multi-modal \yi{information} to \io{a} target user.}
% We also performed a case study to illustrate the effectiveness of our UGT model \iadh{in identifying} useful multi-modal \yi{information} to \io{a} target user.

\balance
\bibliographystyle{ACM-Reference-Format}
\bibliography{reference}

\end{document}